\documentclass[twoside,11pt]{article}

\newif\ifarxiv
\arxivtrue
\ifdefined\forcearxiv\arxivtrue\fi
\ifarxiv\PassOptionsToPackage{preprint}{jmlr2e}\fi
\usepackage{jmlr2e}

\usepackage{amsmath}%
\usepackage{listings}%
\usepackage{xcolor}%
\usepackage{booktabs}%
\usepackage{pifont}%
\usepackage{tikz}%
\usepackage{placeins}%
\usepackage[acronym,nomain,nonumberlist]{glossaries-extra}
\usetikzlibrary{positioning,arrows.meta,shapes.geometric,calc,fit,backgrounds}

\newcommand{\code}[1]{\texttt{\small #1}}
\newcommand{\lmj}{LocoMuJoCo}

\setabbreviationstyle[acronym]{long-short}
\newacronym{il}{IL}{imitation learning}
\newacronym{rl}{RL}{reinforcement learning}
\glsdisablehyper

\lstdefinestyle{pylmj}{%
  language=Python,
  basicstyle=\ttfamily\footnotesize,
  keywordstyle=\color{blue!70!black}\bfseries,
  commentstyle=\color{black!50}\itshape,
  stringstyle=\color{purple!70!black},
  frame=single,
  rulecolor=\color{black!30},
  framesep=4pt,
  framerule=0.4pt,
  xleftmargin=4pt,
  xrightmargin=4pt,
  showstringspaces=false,
  breaklines=true,
  columns=fullflexible,
  keepspaces=true,
}

\usepackage{lastpage}
\jmlrheading{XX}{\ifarxiv 2026\else YYYY\fi}{1-\pageref{LastPage}}{MM/YY; Revised MM/YY}{MM/YY}{XX-XXXX}{Firas Al-Hafez, Davide Tateo, Oleg Kaidanov, Aditya Bhatt, Artur Zelik, Zhengyi Luo, and Jan Peters}

\ShortHeadings{LocoMuJoCo}{Al-Hafez et al.}
\firstpageno{1}

\begin{document}

\title{LocoMuJoCo: A Comprehensive Imitation Learning\\Benchmark for Whole-Body Control}

\author{%
  \name Firas Al-Hafez \email firas.al-hafez@tu-darmstadt.de \\
  \name Davide Tateo   \email davide.tateo@tu-darmstadt.de  \\
  \addr Department of Computer Science\\
        Technische Universit\"at Darmstadt\\
        Darmstadt, Hesse, Germany
  \AND
  \name Oleg Kaidanov \email oleg.kaidanov@dfki.de \\
  \name Aditya Bhatt  \email aditya.bhatt@dfki.de  \\
  \name Artur Zelik   \email artur.zelik@dfki.de   \\
  \addr German Research Center for Artificial Intelligence (DFKI)\\
        Darmstadt, Hesse, Germany
  \AND
  \name Zhengyi Luo \email zhengyiluo5@gmail.com \\
  \addr Carnegie Mellon University\\
        Pittsburgh, PA, USA
  \AND
  \name Jan Peters \email peters@ias.tu-darmstadt.de \\
  \addr German Research Center for Artificial Intelligence (DFKI)\\
        Centre for Cognitive Science, Hessian.AI\\
        Technische Universit\"at Darmstadt, Germany%
}

\editor{TBD}

\maketitle

\begin{abstract}%
\lmj{} is an open-source benchmark for \gls{il} and
\gls{rl} on whole-body control. The library couples
MuJoCo (CPU) with MJX and the MjWarp backend (GPU) under a single API,
exposes 9 humanoid, 3 (musculo\nobreakdash-)skeletal, and 4 quadruped
embodiments, and ships a curated native motion-capture dataset for
every humanoid, to which it can also retarget AMASS and LAFAN1.
A modular plug-in architecture makes observations, rewards, control modes,
terrains, and domain randomization fully interchangeable, while
compact JAX implementations of many \gls{rl} and \gls{il} algorithms
provide JIT-compiled end-to-end training, and the companion
\code{loco-mujoco-real} repository provides sim-to-real deployment. We describe the design,
the SMPL-to-robot and robot-to-robot retargeting pipeline, and a usage
example. \lmj{} is released under MIT at
\url{https://github.com/robfiras/loco-mujoco}.
\end{abstract}


\glsreset{il}
\glsreset{rl}

\section{Introduction}
\label{sec:intro}

Humanoid robots have advanced rapidly in recent years, and improvements
in hardware have made the sim\nobreakdash-to\nobreakdash-real transfer of
learned controllers increasingly practical. Because humans and humanoid
robots share broadly similar embodiments, human motion data provide an
abundant source of training demonstrations. Despite this
progress, the \gls{il}/\gls{rl} benchmarking landscape
remains fragmented across single-robot codebases, manipulation-only suites,
and toy locomotion tasks, forcing researchers to re-implement data pipelines,
retargeting methods, simulators, and evaluation metrics whenever they change
embodiments.

The first version of \lmj{}~\citep{alhafez2023workshop} paired realistic
humanoid models with motion-capture datasets and a common evaluation
protocol, but ran on CPU MuJoCo only. Version~1.0 rebuilt the library
around JAX, adding a GPU-parallel MJX backend, the retargeting pipeline,
many new embodiments, and JAX implementations of \gls{rl} and \gls{il}
algorithms. The current version adds an optional MjWarp backend,
further algorithms, and the companion
\code{loco-mujoco-real} sim-to-real deployment pipeline.

\section{Related Work}
\label{sec:related}

\begin{table}[t]
\centering\footnotesize
\renewcommand{\arraystretch}{1.0}
\setlength{\tabcolsep}{3pt}
\begin{tabular}{lccccc}
\toprule
                       & \textbf{Dual backend}      & \textbf{Built-in}        & \textbf{Biomechanics} & \textbf{Trajectory} & \textbf{Sim-to-real}\\
                       & \textbf{(NumPy + JAX)}     & \textbf{retargeting}     & \textbf{(muscles)}    & \textbf{metrics}    & \textbf{tooling}    \\
\midrule
HumanoidBench          & \ding{55}                  & \ding{55}                & \ding{55}             & \ding{55}           & \ding{55}           \\
MyoSuite               & \ding{55}                  & \ding{55}                & \ding{51}             & \ding{55}           & \ding{55}           \\
mjlab                  & \ding{55}                  & \ding{55}                & \ding{55}             & \ding{55}           & \ding{51}           \\
ProtoMotions           & \ding{55}                  & \ding{51}                & \ding{55}             & \ding{55}           & \ding{51}           \\
MimicKit               & \ding{55}                  & \ding{55}                & \ding{55}             & \ding{55}           & \ding{55}           \\
\textbf{\lmj{}}     & \ding{51}                  & \ding{51}                & \ding{51}             & \ding{51}           & \ding{51}           \\
\bottomrule
\end{tabular}
\caption{Comparison with recent humanoid motion-imitation libraries.
\ding{51}/\ding{55} denote built-in support/absence, based on cited
papers and public documentation (2026).}
\label{tab:comparison}
\end{table}

Recent humanoid \gls{il}/\gls{rl} libraries each emphasize one strength:
HumanoidBench~\citep{sferrazza2024humanoidbench} (\gls{rl} tasks),
MyoSuite~\citep{caggiano2022myosuite} (musculoskeletal models),
mjlab~\citep{zakka2026mjlab} (Isaac\nobreakdash-Lab-style API on
MuJoCo-Warp), ProtoMotions~\citep{tessler2024protomotions} (GPU motion
tracking with retargeting), and MimicKit~\citep{peng2025mimickit}
(motion-imitation framework). None combines a single API spanning a
serial NumPy and a parallel JAX backend, automated retargeting onto any
registered robot, biomechanical models, trajectory-similarity metrics,
and a sim-to-real pipeline (Table~\ref{tab:comparison};
Appendix~\ref{app:related}). General-purpose GPU frameworks such as
Brax~\citep{freeman2021brax} and MuJoCo
Playground~\citep{zakka2025playground} share our simulation stack but
provide no imitation-learning layer.

\begin{figure}[t]
\centering
\includegraphics[width=.8\textwidth]{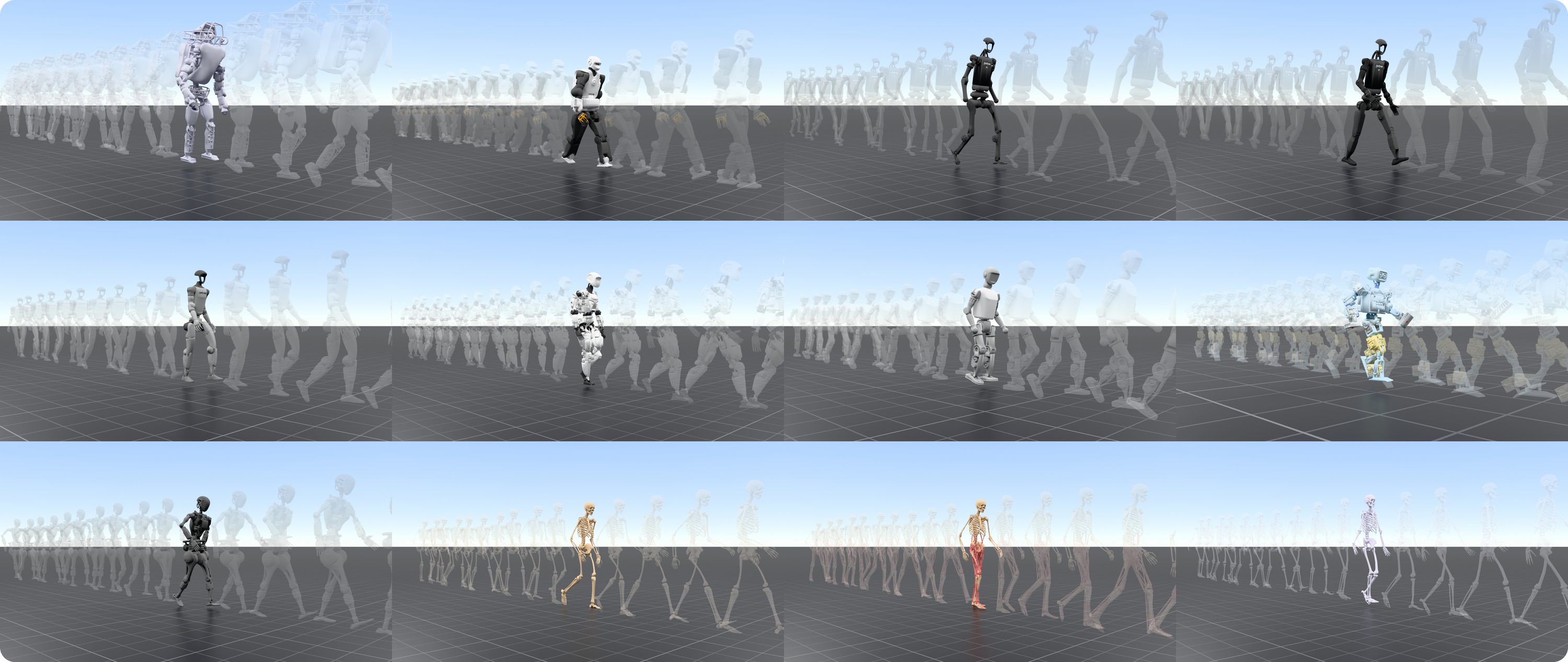}
\caption{The 12 humanoid and (musculo-)skeletal embodiments replaying
one walking trajectory; ghosted poses show time. All share one training,
plug-in, and dataset interface.}
\label{fig:envs}
\end{figure}

\section{The \lmj{} Library}
\label{sec:design}

\subsection{Dual Environment Backends}
\label{sec:backend}
\lmj{} executes each environment through either a serial NumPy/MuJoCo
backend or a vectorized JAX/MJX backend. Environment components---including
observations, rewards, controls, and termination conditions---are implemented
once against a numerical-backend argument. Because \code{jax.numpy} closely
mirrors NumPy, most code is shared verbatim and only backend-specific operations
branch, avoiding duplicate environment logic. \lmj{} can additionally use
MjWarp via \code{use\_mjwarp=True}, which reaches up to $7.3\times10^{5}$
steps per second on one consumer GPU, on par with mjlab and MuJoCo Playground
on their own tasks (Appendix~\ref{app:throughput}).

\subsection{Modular Architecture}
\label{sec:modular}
The simulation core (\code{loco\_mujoco/core}) is decomposed into
interchangeable plug-ins for observations, rewards, control functions
(direct motor or muscle actuation, PD), domain randomization (gravity,
friction, mass, joint dynamics, PD gains, observation noise), terrain
(flat, rough heightfield), and initial/terminal state handlers. Plug-ins
are stateless: each keeps its per-environment state in an explicit Flax
dataclass that is passed through the step, so the entire environment
step is JIT-compilable and nothing lives in hidden global state. The
same training loop therefore runs across robots and across \gls{rl} and
\gls{il} (Appendix~\ref{app:modular}).

\subsection{Embodiments, Datasets, and Retargeting}
\label{sec:data}

\lmj{} ships 9 humanoid robots (Unitree~H1/H1v2/G1, Atlas, Talos,
Apollo, Booster~T1, ToddlerBot, Fourier~GR1T2), 3 (musculo-)skeletal
SMPL-derived human models (\code{MyoSkeleton},
\code{SkeletonTorque}, \code{SkeletonMuscle}), and 4 quadrupeds
(Unitree~A1/Go2, BD~Spot, ANYmal~C). Three dataset sources are integrated: a native, retargeted
\lmj{} dataset hosted on the HuggingFace Hub, the LAFAN1 motion-capture
collection, and the AMASS aggregate (the latter requires the user to
accept the SMPL license). The~companion \code{loco-mujoco-real}
repository adds real-robot environments for the Unitree~G1 and the
AgiBot~X2~Ultra, which also serve as worked examples of adding a new
robot to \lmj{}, including its SMPL retargeting configuration.
The SMPL-to-robot and robot-to-robot procedures are detailed
in Appendix~\ref{app:retargeting}; they map human motion to
robots and transfer it across embodiments without manual
per-trajectory annotation.

\subsection{Algorithms, Training, and Evaluation}
\label{sec:algos}
We provide compact JAX/Flax implementations of
PPO~\citep{schulman2017ppo},
GAIL~\citep{ho2016gail},
a DeepMimic-style motion-tracking algorithm~\citep{peng2018deepmimic},
and the world-model-based SuperMimic~\citep{liao2026supermimic},
together with BPTT-PPO, a Transformer-based history-aware PPO,
SAC~\citep{haarnoja2018sac}, TD3~\citep{fujimoto2018td3},
S2PG~\citep{alhafez2024s2pg}, and a DAgger-style distillation pipeline
for compressing specialist policies into a single multi-task student. Each training
script fuses environment stepping and the policy/critic update into a
single JIT-compiled loop, removing Python overhead between steps.
For evaluation, the metrics module is backed by the JAX-native
\code{metrx} library~\citep{pompetzki2024metrx} and supplies trajectory
similarity measures (Euclidean, dynamic time warping, and the discrete
Fr\'echet distance) over 11 joint, body, and site quantities. Appendix~\ref{app:experiments} reports
the experiments: 715 runs on 11 embodiments.

\paragraph{Usage example.}
\label{sec:usage}
An environment is created with \code{ImitationFactory.make} and stepped
with \code{reset}/\code{step}, or in parallel with
\code{mjx\_reset}/\code{mjx\_step} (full example in
Appendix~\ref{app:usage}).

\subsection{Real-World Deployment}
\label{sec:real}
\begin{figure}[t]
\centering
\includegraphics[width=\textwidth]{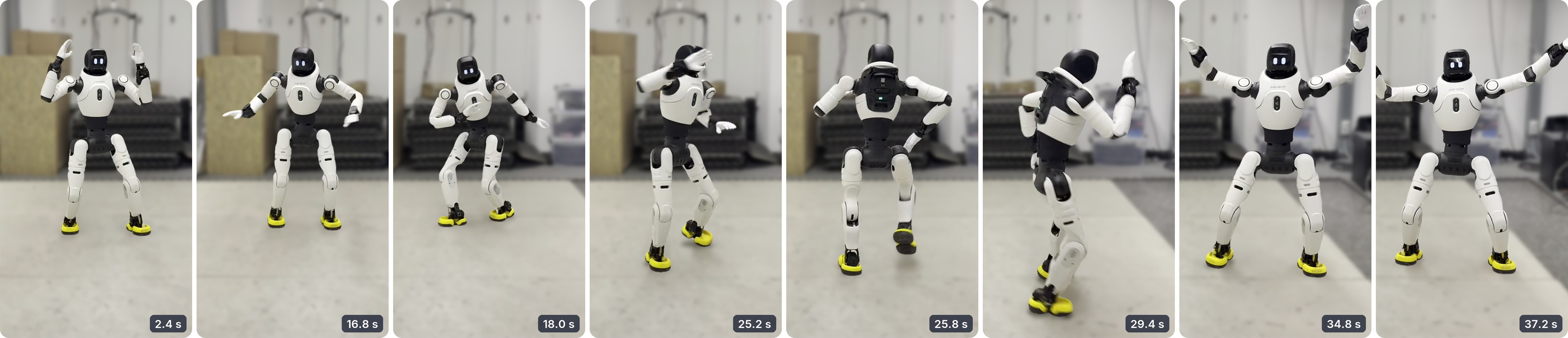}
\caption{Sim-to-real transfer on the AgiBot~X2~Ultra: one motion-tracking
policy trained in \lmj{}, running on the real robot over a 43\,s motion.}
\label{fig:sim2real}
\end{figure}

Policies trained in \lmj{} transfer to hardware through
\code{loco-mujoco-real}, which provides sim-to-real tooling for two
robots, the Unitree~G1 and the AgiBot~X2~Ultra: it adds the robot's
environments, exports the trained policy, and runs it in a sim-to-sim
check and on the robot with the same observation and control code.
Figure~\ref{fig:sim2real} shows a
motion-tracking policy for the AgiBot~X2~Ultra, trained with domain
randomization on the MjWarp backend, executing a 43\,s whole-body motion
at 50\,Hz on the real robot.

\section{Availability}
\label{sec:availability}

{\sloppy \lmj{} is released under MIT at
\url{https://github.com/robfiras/loco-mujoco}, with the companion
sim-to-real pipeline at
\url{https://github.com/robfiras/loco-mujoco-real}, documentation at
\url{https://loco-mujoco.readthedocs.io/}, the robot models at
\url{https://github.com/robfiras/loco-mujoco-models}, and datasets at
\url{https://huggingface.co/datasets/robfiras/loco-mujoco-datasets}.
The test suite covers over 92\,\% of the code
(Appendix~\ref{app:testing}), and contributions are welcome through
GitHub.\par}

\clearpage
\acks{This work was funded by the German Federal Ministry of Research,
Technology and Space (BMFTR) within the Robotics Institute Germany (RIG),
under grant no.~16ME1001 (TU~Darmstadt) and grant no.~16ME1010 (DFKI). It
was also
partially supported by the German Science Foundation (DFG) under grant
number SE1042/41-1 and by the German Federal Ministry of Education and
Research (BMBF) within the collaborative KIARA project (grant
no.~13N16274). The authors declare no competing interests.}

\appendix
\raggedbottom

\section{Detailed Comparison to Related Work}
\label{app:related}

This appendix expands the per-benchmark commentary that the main-text
related work (Section~\ref{sec:related}) summarized in a single
paragraph. We adopt the comparison axes of
Table~\ref{tab:comparison} and add context that the page-budget of the
main text could not accommodate.

\paragraph{HumanoidBench~\citep{sferrazza2024humanoidbench}.}
HumanoidBench is a CPU-MuJoCo benchmark of 27 whole-body manipulation
and locomotion tasks designed around the Unitree~H1 with dexterous
hands and the Agility~Digit lower body. Its primary contribution is a
task-design suite covering long-horizon and contact-rich behavior;
training is \gls{rl}-only and no expert demonstrations are shipped, so the
suite cannot be used for \gls{il} out-of-the-box. The
simulator is the standard CPU MuJoCo binding, with no JAX backend and
no built-in GPU parallelism, which limits throughput at the high seed
counts our benchmark targets.

\paragraph{MyoSuite~\citep{caggiano2022myosuite}.}
MyoSuite ships high-fidelity musculoskeletal models of the human body
including a full muscle-actuated humanoid skeleton. It is a strong
peer for the biomechanics axis of Table~\ref{tab:comparison} — and
indeed \lmj{}'s \code{MjxMyoSkeleton} environment reuses the
MyoSkeleton model under MyoSuite's license — but the suite is \gls{rl}-only,
provides no expert \gls{il} data, no retargeting pipeline, and no GPU/JAX
backend. It also does not address sim-to-real transfer.

\paragraph{mjlab~\citep{zakka2026mjlab}.}
mjlab adopts the manager-based environment interface of Isaac~Lab,
trains with RSL-RL~\citep{schwarke2025rslrl}, and uses MuJoCo-Warp for
GPU-accelerated simulation. The
project ships reference velocity-tracking, motion-imitation, and
manipulation tasks. Compared with \lmj{}, mjlab is GPU-only (no
serial NumPy backend at parity), does not provide a curated
human-mocap dataset suite, does not include automated cross-robot
retargeting, and has no biomechanical models. The environment design
of our companion \code{loco-mujoco-real} repository closely follows
mjlab's G1 motion-tracking setup, which is explicitly positioned for
sim-to-real \gls{rl}.

\paragraph{ProtoMotions~\citep{tessler2024protomotions}.}
ProtoMotions is a GPU-accelerated motion-tracking framework with
built-in human-to-robot retargeting
(PyRoki-based~\citep{kim2025pyroki}) and supports
multiple GPU simulators (IsaacGym, Isaac~Lab, Genesis, NVIDIA Newton).
It demonstrates zero-shot transfer of a single tracking policy onto
the Unitree~G1. The library is closer to a software toolkit than a
benchmark — there is neither a curated, frozen dataset suite nor a
shared evaluation protocol with explicit trajectory-similarity metrics.

\paragraph{MimicKit~\citep{peng2025mimickit}.}
MimicKit unifies three tracking-style algorithms — DeepMimic, AMP, and
ADD — into a single modular \gls{rl} framework targeted at physics-based
character animation. As a methods toolkit it is largely complementary
to a benchmark such as \lmj{}; the differences relevant to
Table~\ref{tab:comparison} are that MimicKit ships no curated robot or
dataset suite, no biomechanical models, no built-in trajectory metrics,
and no sim-to-real instrumentation, and it is not designed for the
parallel-JAX simulation regime that drives \lmj{}'s training
throughput.

\paragraph{General-purpose GPU frameworks.}
Brax~\citep{freeman2021brax} and MuJoCo
Playground~\citep{zakka2025playground} share \lmj{}'s simulation
stack but are not imitation-learning libraries, which is why they are
not part of Table~\ref{tab:comparison}. Brax provides JAX physics
pipelines and \gls{rl} training for a set of continuous-control tasks.
The original release of MuJoCo Playground is a suite of MJX
environments for control, locomotion, and manipulation, trained in JAX
and deployed on several robots; its current version moved to PyTorch
and MuJoCo~Warp and adds a whole-body motion-tracking task
for the Unitree~G1, which loads motions retargeted by external tools.
Neither provides motion datasets or a retargeting pipeline of its own.
\lmj{} builds on MJX and MuJoCo~Warp as well and adds the
imitation-specific layer on top: datasets for every embodiment,
retargeting, imitation algorithms, and trajectory-similarity metrics.

\paragraph{Summary.}
The picks above are the libraries researchers in humanoid \gls{il}/\gls{rl} today
benchmark against. They are individually strong on one or two of the
five axes in Table~\ref{tab:comparison}, but none simultaneously
covers the combination of dual-backend simulation, automated mocap
retargeting onto any registered robot, biomechanical models,
JAX-native trajectory metrics, and the companion \code{loco-mujoco-real}
deployment pipeline that \lmj{} provides.

\paragraph{Retargeting pipelines.}
Our retargeting builds on the SMPL-fitting approach of PHC and
H2O~\citep{luo2023phc,he2024h2o}, which targets individual robots, and
extends it into a general, configuration-driven pipeline: adding a
robot only requires a configuration file and 15 mimic sites in its
model, and trajectories transfer
between robots. To the best of our knowledge, it was the first
open-source pipeline of this generality, introduced in \lmj{}
v1.0.1;\footnote{\url{https://github.com/robfiras/loco-mujoco/releases/tag/v1.0.1}}
earlier code, including ProtoMotions~\citep{tessler2024protomotions},
hard-coded the mappings for one or two robots.
Later work
builds on similar ideas: GMR~\citep{araujo2026gmr} uses non-uniform
scaling and differential inverse kinematics instead of a shape fit,
targeting real-time use and fewer artifacts, while
OmniRetarget~\citep{yang2025omniretarget} preserves contacts with
objects and terrain. \lmj{} instead fits SMPL to each robot once,
tracks the resulting $SE(3)$ targets with soft MuJoCo constraints, and
reuses the same mapping for robot-to-robot transfer.

\paragraph{Foundational components.}
\lmj{} builds on several open-source software and data assets
without which the library could not exist. The simulation backbone is
the MuJoCo physics engine~\citep{todorov2012mujoco} and its
JAX-vectorized counterpart MJX. The retargeting pipeline targets the
SMPL parametric human body model~\citep{loper2015smpl}, with
human-mocap data drawn from the AMASS unified
corpus~\citep{mahmood2019amass} and the Ubisoft LaForge LAFAN1
animation dataset~\citep{harvey2020lafan1}. AMASS clips are retargeted on demand under
the user's accepted SMPL license. Trajectory-similarity metrics are
provided by the JAX-native \code{metrx}
library~\citep{pompetzki2024metrx}.

\section{Modular Architecture}
\label{app:modular}

The headline architectural decision of \lmj{} is that everything
that varies across robots, tasks, and research questions is a
plug-in. Section~\ref{sec:modular} lists the slots; this appendix
explains why each one is a slot and how they compose into the two
canonical use cases of the library — pure \gls{rl} for legged
locomotion and \gls{rl}-based \gls{il} from mocap.

\paragraph{Why plug-ins.} Robots in \lmj{} differ in degrees of
freedom (Apollo's hands alone add dozens of DoF over a stripped-down
H1), in actuator type (torque vs.\ PD vs.\ muscle), and in the observation
channels exposed to the policy (e.g.\ proprioception alone vs.\ proprioception with foot
contacts). Hard-coding any of these would force one
\code{Env} class per (robot, task) pair and would block the very
cross-embodiment experiments that motivate the library. Instead,
each environment is parameterized by a small set of orthogonal
components:

\begin{itemize}
\item \textbf{Observations.} A list of named observation specs is
      composed at construction time; each spec returns a fixed-shape
      vector slice (joint positions, joint velocities, base
      orientation, mimic-site positions, contact forces, $\ldots$),
      and the environment concatenates them into the policy input.
      This is what makes \emph{the same training script} work on
      every embodiment despite their different observation
      dimensionalities.
\item \textbf{Rewards.} The shipped reward classes range from
      \code{NoReward} (pure dynamics) and locomotion-style
      shaping (\code{TargetXVelocityReward},
      \code{TargetVelocityGoalReward}, \code{LocomotionReward},
      \code{TargetVelocityTrajReward}) to imitation-style
      \code{MimicReward}, which scores tracking of the
      relative-site quantities described in
      Appendix~\ref{app:experiments}. Switching from velocity
      tracking to motion imitation is a single argument change.
\item \textbf{Domain randomization.} The default randomizer has
      independent switches for gravity, per-geom friction /
      stiffness / damping, link masses, the root center-of-mass
      offset, per-DoF joint friction loss / damping / armature,
      PD gains, and observation noise. The sampled parameters are
      stored in a Flax dataclass, so they live in explicit
      JIT-visible state rather than mutable Python globals.
\item \textbf{Initial-state distribution.} Resetting can use the
      model's default or a user-specified configuration, a uniform
      sample over a loaded mocap clip
      (state-of-the-art for DeepMimic-style training), or from a
      user-defined sampler. The same hook is what lets us reset
      to a particular phase of a clip during evaluation when
      computing trajectory-similarity metrics.
\item \textbf{Termination conditions.} The terminal-state handler
      is selected by name: no termination, a root-height range, a
      root height and orientation range derived from the dataset,
      or a bound on the root's height and tilt error with respect
      to the current reference frame. Pure-\gls{rl} setups
      typically terminate on a fall only; imitation setups also
      terminate when the policy strays too far from the reference
      pose, since this is what produces the truncated rollouts
      visible in the per-clip evaluation metrics.
\item \textbf{Control mode.} \code{DefaultControl}, which passes
      the action directly to the model's actuators (motors or
      muscles), and \code{PDControl} define the action
      space exposed to the policy. The choice changes both the learning
      problem and the action dimensionality without touching the policy
      network or training loop. Torque control is high-bandwidth and stiff.
      PD control supplies lower-bandwidth reference targets and is commonly
      deployed in sim-to-real legged systems, whereas muscle actuation supports
      biomechanics studies on the SMPL-derived skeletal models. The training
      procedure for the muscle-actuated (musculo\nobreakdash-)skeletal
      models follows \citet{geiss2024exciting}.
\item \textbf{Terrain.} Terrain plug-ins (flat, or a rough
      heightfield regenerated at reset) modify the world geometry independently of every
      other slot. This decoupling matters because terrain
      curricula and rough-ground evaluation are routine in
      legged-robot research, and \lmj{} lets the user keep
      everything else fixed — observations, rewards, datasets,
      randomization — while swapping the ground the robot walks on,
      so terrain effects are not confounded with changes to the
      training setup.
\end{itemize}

\paragraph{Why this scales.} Because every plug-in is stateless
and keeps its per-environment state in a Flax dataclass,
the entire environment step compiles cleanly under
\code{jax.jit} and \code{jax.vmap}. A practitioner studying
domain randomization on Booster~T1, an \gls{il}
researcher comparing GAIL and DeepMimic on Apollo, and a
biomechanics-leaning user training muscle policies on
\code{MjxSkeletonMuscle} all share the same training loop and
metric pipeline. \lmj{}'s task factories assemble each robot-task
pair from the selected plug-ins, so these use cases differ only in
configuration. The
experiments in this paper
(Appendix~\ref{app:experiments}) — covering algorithm
and embodiment comparisons — are themselves an example of this
scalability: every cell in every one of them is the same env
core under different plug-in choices.

\paragraph{Embodiment provenance and body scaling.} The Atlas
environment is adapted from \citet{alhafez2023lsiq}. Beyond the rigid
robots, the (musculo\nobreakdash-)skeletal models
(\code{SkeletonTorque}, \code{SkeletonMuscle})
support body scaling: the same skeleton can be instantiated at different
anthropometric sizes without any change to the observation, reward, or
training pipeline — something the fixed-geometry robot embodiments do not
permit. This was implemented to support research on different human
sizes, following \citet{alhafez2024s2pg} and \citet{badie2025bioinspired}.

\section{Usage Example: Full Agent Loop}
\label{app:usage}

Section~\ref{sec:usage} describes the interface in one sentence.
Figure~\ref{alg:usage-full}
shows the full agent loop in both flavours of the library: the CPU loop
on top uses the gymnasium-style synchronous API and is what one
would use during prototyping, debugging, or rendering; the MJX
block below uses \code{jax.jit} and \code{jax.vmap} to step a
batch of \code{n\_envs} environments in parallel on GPU and is
what the algorithm implementations bundled with \lmj{} use under
the hood. The two share the same observation, reward, termination,
and randomization plug-ins (Appendix~\ref{app:modular}); only the
step/reset call sites differ. The environment is constructed once and
shared by both loops.

\begin{figure}[t]
\begin{lstlisting}[style=pylmj]
import numpy as np
import jax
from loco_mujoco import ImitationFactory

# ----- Shared environment (constructed once) --------------------------
env, traj = ImitationFactory.make(
    "MjxUnitreeH1", default_dataset_conf=dict(task="squat"))
action_dim = env.info.action_space.shape[0]

# ----- CPU agent loop --------------------------------------------------
obs = env.reset()
for i in range(1000):
    action = np.random.randn(action_dim)
    obs, reward, absorbing, done, info = env.step(action)
    env.render()
    if absorbing or done:
        obs = env.reset()

# ----- MJX agent loop (JIT + vmap over n_envs) -------------------------
env.reset()
n_envs = 1024
key = jax.random.key(0)
keys = jax.random.split(key, n_envs + 1)
key, env_keys = keys[0], keys[1:]

mjx_reset = jax.jit(jax.vmap(env.mjx_reset,         in_axes=(0, None)))
mjx_step  = jax.jit(jax.vmap(env.mjx_step,          in_axes=(0, 0, None)))
sample    = jax.jit(jax.vmap(env.sample_action_space))

state = mjx_reset(env_keys, traj)
for i in range(1000):
    keys = jax.random.split(key, n_envs + 1)
    key, action_keys = keys[0], keys[1:]
    action = sample(action_keys)
    state = mjx_step(state, action, traj)
\end{lstlisting}
\caption{Full CPU and MJX (JIT + \code{vmap}) agent loops. The environment
is constructed once at the top and shared by both execution paths; only
the step/reset call sites differ.}
\label{alg:usage-full}
\end{figure}

\section{Retargeting Pipeline}
\label{app:retargeting}

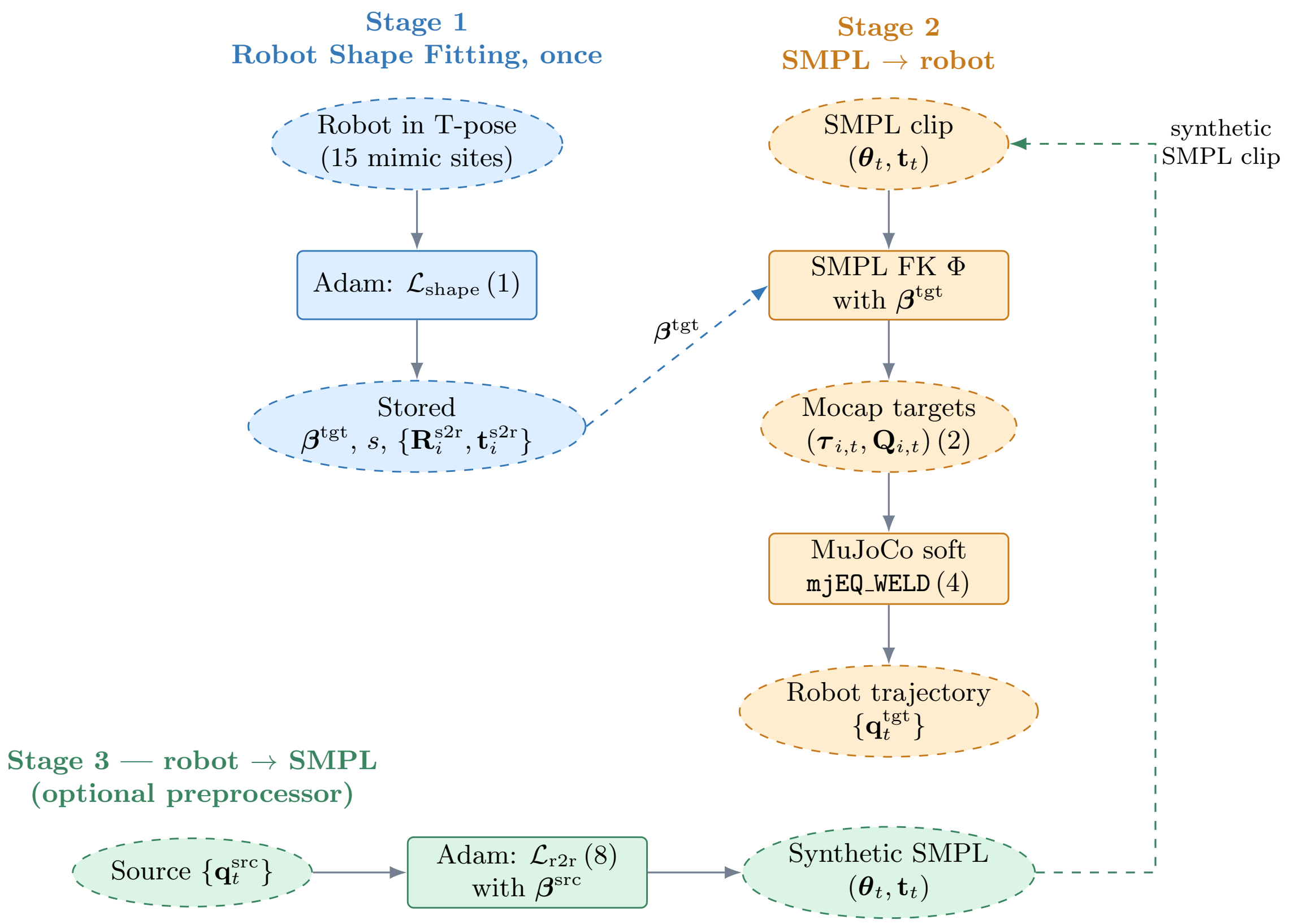
\begin{figure}[t]
\centering
\definecolor{lmjStageOne}{HTML}{DCEEFF}
\definecolor{lmjStageTwo}{HTML}{FFEDCF}
\definecolor{lmjStageThree}{HTML}{DCF4E6}
\definecolor{lmjBlue}{HTML}{3478B8}
\definecolor{lmjOrange}{HTML}{C77A1B}
\definecolor{lmjGreen}{HTML}{398461}
\definecolor{lmjFlow}{HTML}{738091}
\begin{tikzpicture}[
  node distance=7mm and 11mm,
  >=Latex,
  every node/.style={font=\footnotesize},
  block/.style={rectangle, rounded corners=2pt, align=center,
                minimum height=8mm, minimum width=28mm, inner sep=2pt,
                line width=.55pt},
  data/.style ={ellipse, dashed, align=center,
                minimum height=8mm, minimum width=28mm, inner sep=1pt,
                line width=.55pt},
  stageone/.style={draw=lmjBlue, fill=lmjStageOne},
  stagetwo/.style={draw=lmjOrange, fill=lmjStageTwo},
  stagethree/.style={draw=lmjGreen, fill=lmjStageThree},
  grouplab/.style={font=\bfseries\footnotesize},
  flow/.style={->, draw=lmjFlow, semithick},
  cacheflow/.style={->, dashed, draw=lmjBlue, semithick}
]

\node[data, stageone] (s1a) {Robot in T-pose\\(15 mimic sites)};
\node[block, stageone, below=of s1a] (s1b) {Adam: $\mathcal{L}_\text{shape}$\,\eqref{eq:shape}};
\node[data, stageone, below=of s1b] (s1c) {Stored\\$\boldsymbol{\beta}^{\text{tgt}},\, s,\, \{\mathbf{R}^{\text{s2r}}_i, \mathbf{t}^{\text{s2r}}_i\}$};
\node[grouplab, text=lmjBlue, above=2mm of s1a, align=center] {Stage 1\\Robot Shape Fitting, once};
\draw[flow] (s1a) -- (s1b);
\draw[flow] (s1b) -- (s1c);

\node[data, stagetwo, right=24mm of s1a] (s2a) {SMPL clip\\$(\boldsymbol{\theta}_t, \mathbf{t}_t)$};
\node[block, stagetwo, below=of s2a] (s2b) {SMPL FK $\Phi$\\with $\boldsymbol{\beta}^{\text{tgt}}$};
\node[data, stagetwo, below=of s2b] (s2c) {Mocap targets\\$(\boldsymbol{\tau}_{i,t}, \mathbf{Q}_{i,t})$\,\eqref{eq:target-pos}};
\node[block, stagetwo, below=of s2c] (s2d) {MuJoCo soft\\\texttt{mjEQ\_WELD}\,\eqref{eq:mj-step}};
\node[data, stagetwo, below=of s2d] (s2e) {Robot trajectory\\$\{\mathbf{q}^{\text{tgt}}_t\}$};
\node[grouplab, text=lmjOrange, above=2mm of s2a, align=center] {Stage 2\\SMPL $\to$ robot};
\draw[flow] (s2a) -- (s2b);
\draw[flow] (s2b) -- (s2c);
\draw[flow] (s2c) -- (s2d);
\draw[flow] (s2d) -- (s2e);

\node[data, stagethree, below=8mm of s2e] (s3c) {Synthetic SMPL\\$(\boldsymbol{\theta}_t,\mathbf{t}_t)$};
\node[block, stagethree, left=of s3c] (s3b) {Adam: $\mathcal{L}_\text{r2r}$\,\eqref{eq:r2r-loss}\\with $\boldsymbol{\beta}^{\text{src}}$};
\node[data, stagethree, left=of s3b] (s3a) {Source $\{\mathbf{q}^{\text{src}}_t\}$};
\node[grouplab, text=lmjGreen, above=2mm of s3a, align=center] {Stage 3 — robot $\to$ SMPL\\(optional preprocessor)};
\draw[flow] (s3a) -- (s3b);
\draw[flow] (s3b) -- (s3c);

\draw[cacheflow] (s1c.east) -- (s2b.west)
  node[midway, above, font=\scriptsize] {$\boldsymbol{\beta}^{\text{tgt}}$};
\draw[->, dashed, draw=lmjGreen, semithick] (s3c.east) -- ++(14mm,0) |- (s2a.east)
  node[pos=0.55, right, xshift=3pt, font=\scriptsize, align=center] {synthetic\\SMPL clip};
\end{tikzpicture}
\caption{Three-stage retargeting pipeline. Stage~1 fits a per-robot
SMPL shape $\boldsymbol{\beta}^{\text{tgt}}$ and per-site rigid offsets in a
canonical T-pose (run once per robot, equation~\eqref{eq:shape}).
Stage~2 turns any SMPL motion clip into a target-robot trajectory by
stepping MuJoCo with soft \texttt{mjEQ\_WELD} equality constraints
between mocap targets and mimic sites
(equation~\eqref{eq:mj-step}); SMPL forward kinematics here uses the
target robot's own $\boldsymbol{\beta}^{\text{tgt}}$. Contacts and joint limits
are enforced jointly with constraint satisfaction, which prevents the
ground-penetrating poses that rigid-IK retargeters routinely produce.
Stage~3 (optional) maps a source-robot trajectory back into SMPL
coordinates via gradient descent on equation~\eqref{eq:r2r-loss}
using the \emph{source} robot's $\boldsymbol{\beta}^{\text{src}}$;
the resulting synthetic SMPL clip is then fed into Stage~2 with the
target robot's $\boldsymbol{\beta}^{\text{tgt}}$.}
\label{fig:retargeting}
\end{figure}

This appendix details the three-stage motion-retargeting pipeline
sketched in Section~\ref{sec:data}. It builds on the SMPL-fitting
approach of H2O~\citep{he2024h2o} and generalizes it to any registered
robot. The pipeline transfers human
motion-capture data — represented in the SMPL parametric body
model~\citep{loper2015smpl} — onto any registered humanoid robot, and
additionally supports robot-to-robot motion transfer via the same SMPL
intermediate. Figure~\ref{fig:retargeting} shows the data flow.
Figure~\ref{fig:retargeting-rendered} complements it with rendered
examples of the transformations performed at each stage.

\paragraph{Notation.}
We treat SMPL forward kinematics as a map
$\Phi: \mathbb{R}^{3 N_J} \times \mathbb{R}^{16} \times \mathbb{R}^{3}
\to \mathbb{R}^{N_J \times 3} \times \mathbb{R}^{N_J \times 3 \times 3}$
that takes a per-joint axis-angle pose
$\boldsymbol{\theta}=(\theta_0,\dots,\theta_{N_J-1})$, a body shape
$\boldsymbol{\beta}\in\mathbb{R}^{16}$ and a root translation
$\mathbf{t}\in\mathbb{R}^{3}$, and returns the global joint positions
$\mathbf{J}(\boldsymbol{\theta},\boldsymbol{\beta},\mathbf{t})\in\mathbb{R}^{N_J\times 3}$
and joint rotation matrices
$\mathbf{R}(\boldsymbol{\theta},\boldsymbol{\beta})\in\mathbb{R}^{N_J\times 3\times 3}$.
We use SMPL-H ($N_J{=}52$, including hand chains) to consume
the AMASS corpus, but only the 15 body joints participate in
retargeting. Each registered robot is annotated with 15 \emph{mimic
sites}, each mapped to one SMPL body joint
(Table~\ref{tab:mimic-sites}). We index the 15 SMPL body joints by
$\pi:\{1,\dots,15\}\to\{0,\dots,N_J-1\}$ and the corresponding robot
mimic sites by their MuJoCo site IDs.

\begin{table}[t]
\centering\small
\setlength{\tabcolsep}{6pt}
\renewcommand{\arraystretch}{1.05}
\begin{tabular}{ll|ll}
\toprule
mimic site & SMPL joint & mimic site & SMPL joint\\
\midrule
\code{pelvis\_mimic}      & \code{Pelvis}     & \code{left\_shoulder\_mimic}  & \code{L\_Shoulder} \\
\code{upper\_body\_mimic} & \code{Chest}      & \code{left\_elbow\_mimic}     & \code{L\_Elbow}    \\
\code{head\_mimic}        & \code{Head}       & \code{left\_hand\_mimic}      & \code{L\_Wrist}    \\
\code{left\_hip\_mimic}   & \code{L\_Hip}     & \code{right\_shoulder\_mimic} & \code{R\_Shoulder} \\
\code{left\_knee\_mimic}  & \code{L\_Knee}    & \code{right\_elbow\_mimic}    & \code{R\_Elbow}    \\
\code{left\_foot\_mimic}  & \code{L\_Ankle}   & \code{right\_hand\_mimic}     & \code{R\_Wrist}    \\
\code{right\_hip\_mimic}  & \code{R\_Hip}     & \code{right\_foot\_mimic}     & \code{R\_Ankle}    \\
\code{right\_knee\_mimic} & \code{R\_Knee}    &                               &                    \\
\bottomrule
\end{tabular}
\caption{The 15 mimic-site / SMPL-joint correspondences, defined
once in \texttt{loco\_mujoco/smpl/robot\_confs/defaults.yaml} and
reused unchanged across all registered humanoid robots; per-robot
files set the T-pose joint offsets and, where needed, per-site
constraint parameters.}
\label{tab:mimic-sites}
\end{table}

\subsection{Stage 1: per-robot SMPL shape fitting}
\label{app:retargeting:stage1}

\begin{figure}[t]
\centering
\includegraphics[width=\textwidth]{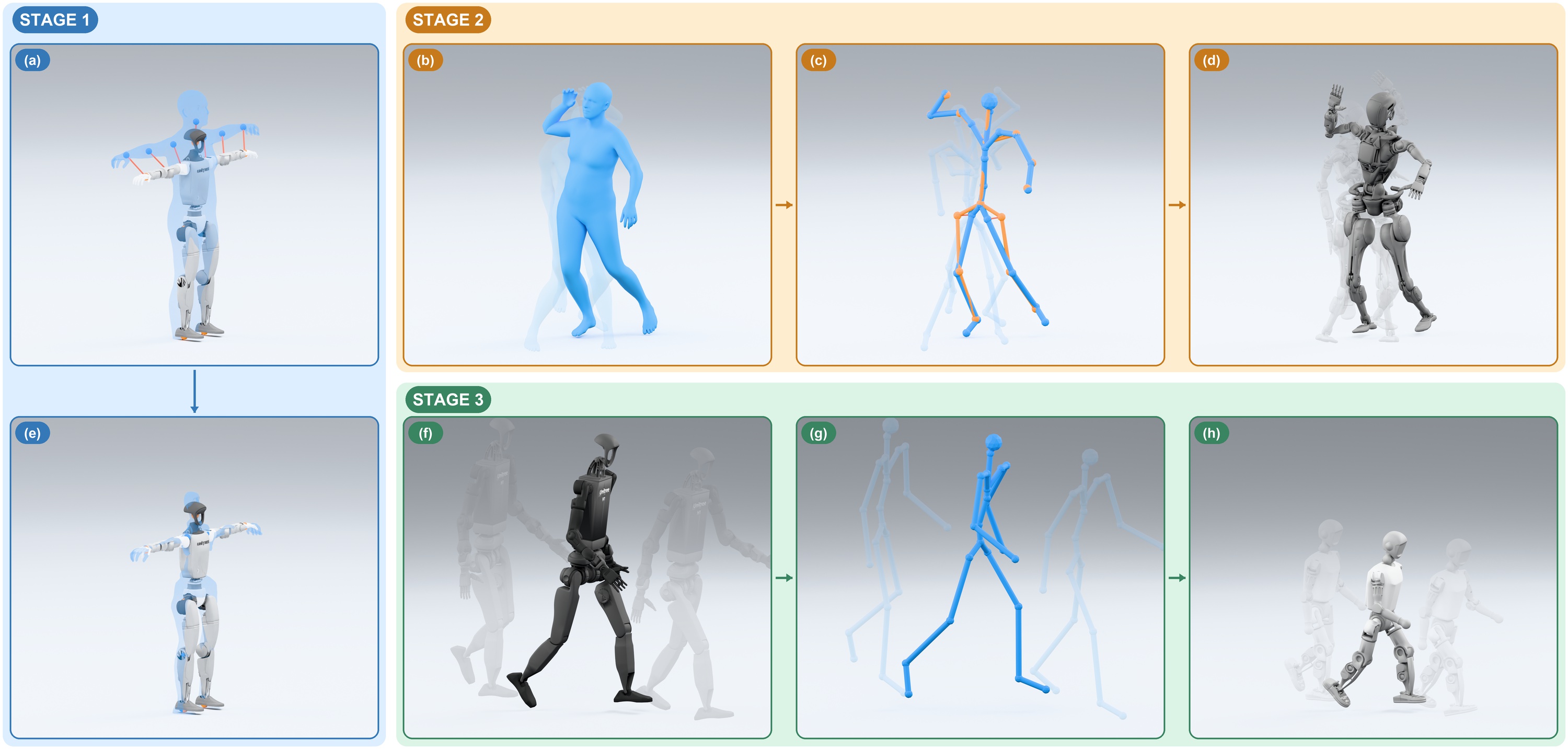}
\caption{Rendered examples of the three retargeting stages. Panels
(a,e) show Stage~1 shape fitting for the Unitree~G1 before and after
optimization; the mean 15-site correspondence error decreases from
$112.1$ to $21.8$~mm. Panels (b--d) show Stage~2 on a dance sequence:
the original SMPL surface, the target-shaped SMPL skeleton (blue) and
15-site target skeleton (orange), and the resulting Fourier~GR1T2
trajectory. Panels (f--h) show Stage~3 on a different walking sequence:
the Unitree~H1v2 source trajectory, its H1v2-shaped synthetic SMPL
intermediate, and the resulting Booster~T1 trajectory. Pale silhouettes
show poses $0.5$~s before and after the highlighted frame; border colors
match the three stages in Figure~\ref{fig:retargeting}.}
\label{fig:retargeting-rendered}
\end{figure}

The robot is posed in a canonical T-pose using the per-joint offsets
of its configuration file, a forward simulation step yields the global mimic-site
positions $\{\mathbf{p}^{\text{robot}}_i\}_{i=1}^{15}$, and we fit an
SMPL shape $\boldsymbol{\beta}$ together with an isotropic body scale
$s\in\mathbb{R}_{>0}$ that align the SMPL body in T-pose with the
robot (the first column of Figure~\ref{fig:retargeting-rendered} shows
the alignment before and after optimization for the Unitree~G1):
\begin{equation}
\min_{\boldsymbol{\beta},\,s}\
\mathcal{L}_\text{shape}(\boldsymbol{\beta},s)
\;=\;\frac{1}{15}\sum_{i=1}^{15}
\bigl\| \mathbf{p}^{\text{robot}}_i \;-\;
        \mathbf{p}^{\text{smpl}}_{\pi(i)}(\boldsymbol{\theta}_T,\boldsymbol{\beta},s) \bigr\|_2,
\label{eq:shape}
\end{equation}

where
$\mathbf{p}^{\text{smpl}}_j(\boldsymbol{\theta}_T,\boldsymbol{\beta},s)
= s\,(\mathbf{J}_j(\boldsymbol{\theta}_T,\boldsymbol{\beta},\mathbf{0}) - \mathbf{J}_0)
+ \mathbf{p}_\text{root}$
is the rescaled SMPL global joint position relative to the pelvis,
$\boldsymbol{\theta}_T$ is the SMPL T-pose, and $\mathbf{p}_\text{root}$
is a fixed root offset. We optimize with Adam for 1000 iterations at
learning rate 0.05 (\texttt{shape\_iterations}, \texttt{shape\_lr} in
\texttt{defaults.yaml}). After convergence, for each site $i$ we also
record a rigid offset
$(\mathbf{R}^{\text{s2r}}_i,\mathbf{t}^{\text{s2r}}_i)\in SE(3)$
aligning the SMPL joint frame to the robot site frame, a feet-height
correction $\Delta z$, and a head-to-feet height ratio. These are
serialized together with $(\boldsymbol{\beta},s)$ to a per-robot
\texttt{shape\_optimized.pkl}.

\subsection{Stage 2: SMPL motion to robot trajectory}
\label{app:retargeting:stage2}

The upper row of Figure~\ref{fig:retargeting-rendered} visualizes this
forward path from an original SMPL motion, through the target-shaped
SMPL skeleton and its 15 target frames, to the resulting robot motion.
Given an SMPL motion sequence
$\{(\boldsymbol{\theta}_t,\mathbf{t}_t)\}_{t=0}^{T-1}$ — for example
an AMASS clip — we evaluate forward kinematics with the fitted shape,
$(\mathbf{J}_t,\mathbf{R}_t)=\Phi(\boldsymbol{\theta}_t,\boldsymbol{\beta},\mathbf{t}_t)$,
and apply the per-site offset to obtain the per-frame target pose of
each mimic site in world coordinates:
\begin{align}
\boldsymbol{\tau}_{i,t} &=
\mathbf{J}_{\pi(i),t}\;-\;\mathbf{R}_{\pi(i),t}\,\mathbf{R}^{\text{s2r}}_i\,\mathbf{t}^{\text{s2r}}_i,
\label{eq:target-pos}\\
\mathbf{Q}_{i,t} &= \mathbf{R}_{\pi(i),t}\,\mathbf{R}^{\text{s2r}}_i.
\label{eq:target-rot}
\end{align}
Two design choices carry the pipeline.
First, the per-site offset $(\mathbf{R}^{\text{s2r}}_i, \mathbf{t}^{\text{s2r}}_i)$
— fitted once in Stage~1 and reused for every frame and every clip
— absorbs the systematic mismatch between SMPL joint centers (which
sit inside the body) and each robot's mimic sites (typically anchored
on the surface), so the per-frame targets are anatomically correct
for the robot rather than for SMPL. Second, providing both a target
translation $\boldsymbol{\tau}_{i,t}$ and a target rotation
$\mathbf{Q}_{i,t}$ lets us use MuJoCo's full $SE(3)$
\texttt{mjEQ\_WELD} constraint instead of the position-only
\texttt{mjEQ\_CONNECT}, which is necessary to track
orientation-sensitive degrees of freedom (e.g.\ wrist twist, pelvis
yaw) and admits the closed-form inverse that Stage~3 reuses.

For each frame, $(\boldsymbol{\tau}_{i,t},\mathbf{Q}_{i,t})$ is loaded
into a MuJoCo mocap body and connected to the corresponding robot site
by an \texttt{mjEQ\_WELD} equality constraint. Crucially the constraint
is \emph{soft}: MuJoCo's solver uses parameters \texttt{solref} and
\texttt{solimp} to express a damped spring-like impedance that admits
finite constraint violation. Per-frame generalized coordinates are then
obtained by warm-starting from the previous frame and integrating
$N_\text{sub}$ MuJoCo sub-steps,
\begin{equation}
\mathbf{q}_t \;=\;
\texttt{mj\_step}^{N_\text{sub}}\!\bigl(
\mathbf{q}_{t-1};\ \mathcal{M}_{\text{robot}},\ \{\boldsymbol{\tau}_{i,t}, \mathbf{Q}_{i,t}\}_{i=1}^{15}\bigr),
\label{eq:mj-step}
\end{equation}
with $N_\text{sub}=25$ by default (\texttt{motion\_iterations}).
Because the same physics step also enforces collisions, joint limits,
and (optionally) actuator-torque limits, the resulting trajectory is more likely to be
physically feasible: the soft constraint allows the optimizer to
favor kinematic plausibility over exact site matching when the two
conflict (for instance, when the human reference would force the foot
through the floor). This contrasts with rigid-IK retargeters, which
can produce ground-penetrating or self-colliding poses.

After processing every frame, if foot geoms are configured for the
robot (none by default), a single per-trajectory floor offset
\begin{equation}
\Delta z_\text{traj} \;=\;
\min\Bigl(\;\texttt{max\_height\_adjustment},\;
          \min_{t,\,g\in\mathcal{G}_\text{feet}}
          \texttt{dist}\bigl(g,\,\texttt{floor}\bigr)\Bigr)
\label{eq:floor-offset}
\end{equation}
is subtracted uniformly from the root height, where
$\mathcal{G}_\text{feet}$ is a per-robot list of foot geoms. This makes
the worst frame just barely contact the floor without altering the
relative motion. Generalized velocities are recovered by central
differences,
$\mathbf{v}_t = (2\Delta t)^{-1}(\mathbf{q}_{t+1}-\mathbf{q}_{t-1})$.

\subsection{Stage 3: robot-to-robot retargeting}
\label{app:retargeting:stage3}

To transfer a trajectory $\{\mathbf{q}^{\text{src}}_t\}_{t=0}^{T-1}$
from a \emph{source} robot onto a \emph{target} robot, we first invert
the source motion back into SMPL coordinates and then route through
Stage~2 with the target robot's shape parameters.
The lower row of Figure~\ref{fig:retargeting-rendered} visualizes this
composition for an H1v2 walking trajectory and a Booster~T1 target;
the blue skeleton is the synthetic SMPL intermediate shared by the two
procedures.

\paragraph{Inverting source $\to$ SMPL.}
We run forward kinematics on the source robot to obtain world-frame
mimic-site poses $(\mathbf{p}^{\text{src}}_{i,t},\mathbf{R}^{\text{src}}_{i,t})$
and remove the source robot's per-site rigid offset to express them
in the SMPL frame:
\begin{align}
\hat{\mathbf{p}}^{\text{src}}_{i,t} &= \mathbf{p}^{\text{src}}_{i,t}\;+\;
\mathbf{R}^{\text{src}}_{i,t}\,\mathbf{t}^{\text{s2r,src}}_i,
\label{eq:src-inv-pos}\\
\hat{\mathbf{R}}^{\text{src}}_{i,t} &=
\mathbf{R}^{\text{src}}_{i,t}\,(\mathbf{R}^{\text{s2r,src}}_i)^{-1}.
\label{eq:src-inv-rot}
\end{align}
We then optimize an SMPL pose-and-translation trajectory
$\{(\boldsymbol{\theta}_t,\mathbf{t}_t)\}_{t=0}^{T-1}$ to fit these
targets while holding the \emph{source} robot's shape
$\boldsymbol{\beta}^{\text{src}}$ fixed:
\begin{equation}
\mathcal{L}_\text{r2r}
\;=\;\frac{\lambda_p}{15T}\!\sum_{i,t}\!
\bigl\|\hat{\mathbf{p}}^{\text{src}}_{i,t}
       - \mathbf{J}_{\pi(i),t}(\boldsymbol{\theta}_t,\boldsymbol{\beta}^{\text{src}},\mathbf{t}_t)\bigr\|_2
\;+\;\frac{\lambda_r}{15T}\!\sum_{i,t}\!
\arccos\!\Bigl(\tfrac{\operatorname{tr}\!\bigl(\mathbf{R}_{\pi(i),t}^{\!\top}\hat{\mathbf{R}}^{\text{src}}_{i,t}\bigr)-1}{2}\Bigr),
\label{eq:r2r-loss}
\end{equation}
with default weights $\lambda_p{=}1.0$ and $\lambda_r{=}0.1$. The
position term uses Euclidean distance between SMPL global joints and
source-frame targets; the rotation term is the geodesic angle between
their orientations. We optimize with Adam for 400 iterations
(\texttt{pose\_iterations}, \texttt{pose\_lr}=0.005), and apply a
1-D Gaussian smoothing kernel on the body-pose component of
$\boldsymbol{\theta}$ each step to suppress jitter
($\texttt{kernel\_size}{=}3,\ \sigma{=}0.5$).

The output is a synthetic SMPL motion clip
$\{(\boldsymbol{\theta}_t,\mathbf{t}_t)\}$ that is dynamically
equivalent — at the level of the 15 correspondences — to the source
robot's motion. We then route it through Stage~2 using the
\emph{target} robot's $\boldsymbol{\beta}^{\text{tgt}}$ to obtain the
target trajectory $\{\mathbf{q}^{\text{tgt}}_t\}$. 

\section{Algorithms}
\label{app:algorithms}

\lmj{} ships a suite of compact JAX/Flax learning algorithms
spanning pure \gls{rl} and \gls{il}.
Each algorithm is implemented in one module, \code{<algo>\_jax.py},
in the package \code{loco\_mujoco/algorithms/} (the additional
algorithms are in its \code{experimental} subpackage) on top of a
small set of shared networks and base classes, with
the training loop and policy update fused into a single JIT-compiled
function built from \texttt{lax.scan} so that environment stepping, advantage computation,
and gradient updates execute on-device without Python overhead. The
following short summaries describe what each algorithm does and the
role it plays in the benchmark.

\subsection{Core algorithms}
\label{app:algorithms:core}

\paragraph{PPO~\citep{schulman2017ppo}.}
\emph{Proximal Policy Optimization} is an on-policy actor-critic
policy-gradient method that improves the policy by maximizing a
clipped surrogate objective, bounding how much the policy can shift
per update. PPO is the default pure-\gls{rl} backbone in \lmj{} and is
also reused inside the imitation algorithms below as the policy
optimizer when they need an inner-loop policy update. It also trains
data-parallel on several GPUs: each device steps an equal share of the
environments and the gradients are averaged across devices, as in the
training script of \code{loco-mujoco-real}.

\paragraph{GAIL~\citep{ho2016gail}.}
\emph{Generative Adversarial Imitation Learning} trains a
discriminator to distinguish expert state(-action) pairs from
policy-generated ones and uses the log-discriminator output as the
reward signal for the policy (optimized by PPO in our implementation).
GAIL is useful in \gls{il} settings where expert demonstrations are
available but a hand-crafted task reward is not, and where exact
frame-by-frame tracking is unnecessary.

\paragraph{AMP~\citep{peng2021amp}.}
\emph{Adversarial Motion Priors} combines a GAIL-style discriminator
on \emph{state transitions} (encoding motion style) with an additive
user-supplied task reward (e.g.\ velocity tracking). The agent learns
to pursue the task while staying within the distribution of stylistic
motion shown in the reference dataset. AMP is the recommended
algorithm for stylized \gls{rl} on \lmj{}'s humanoid envs.

\paragraph{DeepMimic~\citep{peng2018deepmimic}.}
\emph{DeepMimic} is a tracking-style \gls{il} method that rewards the
policy, at every step, for staying close to a reference motion clip.
\lmj{} trains it with PPO and the \code{MimicReward} class, a
DeepMimic-style reward over joint positions and velocities and the
root-relative positions, orientations, and velocities of the mimic
sites; it is the default \gls{il} training signal for
the curated humanoid mocap clips.

\paragraph{SuperMimic~\citep{liao2026supermimic}.}
A model-based motion-tracking method: a world model is fitted to
simulator transitions, and the deterministic policy is improved by
backpropagating through short imagined rollouts of that model, with a
TD($\lambda$) critic bootstrapping the tail. In our implementation an
imagined step is an ordinary environment step in which the learned
dynamics replace the simulator, so observations, rewards, and
terminations are the environment's own functions and the algorithm is
independent of the robot, the observation, and the reward.
Appendix~\ref{app:exp4} reports results on the
two real-robot environments.

\subsection{Additional algorithms}
\label{app:algorithms:additional}

\paragraph{SAC~\citep{haarnoja2018sac}.}
\emph{Soft Actor-Critic} is an off-policy maximum-entropy
actor-critic that simultaneously maximizes expected return and policy
entropy. It is typically more sample-efficient than PPO on continuous
control and is offered as an off-policy alternative for users with a
tighter sample budget.

\paragraph{TD3~\citep{fujimoto2018td3}.}
\emph{Twin Delayed Deep Deterministic Policy Gradient} is an
off-policy deterministic actor-critic that mitigates value-function
over-estimation via twin critics, target-policy smoothing, and
delayed policy updates. Provided as a deterministic-policy
counterpoint to SAC.

\paragraph{S2PG~\citep{alhafez2024s2pg}.}
\emph{Stochastic Stateful Policy Gradient} splits a stateful policy
into a stochastic policy-state transition kernel and a stateless
policy, and trains both without backpropagation through time.
This avoids the gradient instability of full BPTT while still
letting the policy exploit memory.
Useful for partially-observable variants of the locomotion tasks.

\paragraph{BPTT-PPO.}
A PPO variant that backpropagates the policy-gradient signal through
an unrolled time horizon, supporting fully recurrent (LSTM/GRU)
policies. The training loop unrolls $T_\text{bptt}$ steps in
\texttt{lax.scan} per gradient step. Useful when the task genuinely
requires long-range memory.

\paragraph{History-aware PPO.}
A PPO variant whose observation is augmented with a fixed-length
stack of past observations and actions, encoded either by an MLP or
by a Transformer block (multi-head self-attention with layer norm).
This exposes short-horizon context to the policy without the
optimization cost of full recurrence and is a pragmatic middle
ground between standard PPO and BPTT-PPO when memory needs are
moderate.

\paragraph{DAgger Distillation.}
\emph{Dataset Aggregation} implements teacher–student
distillation~\citep{ross2011dagger}.
At rollout time each parallel env independently picks the student or
the teacher with probability $\beta$ (sticky for a random number of
steps), but every transition is labeled with the \emph{teacher's}
action on the visited observation. The student is then trained
supervised — minibatch NLL of the student Gaussian at the stored
teacher action. This is \lmj{}'s default distillation pipeline,
used to compress per-clip or per-embodiment PPO specialists into a
single multi-task generalist.

\section{Testing and Continuous Integration}
\label{app:testing}

\lmj{} ships a \code{pytest} suite of about 700 tests that covers the
environments on both backends, the observation, reward, control,
domain-randomization, and termination components, the trajectory and dataset
pipeline, the retargeting pipeline, the rendering and export tools, and a short
training run of every included algorithm. The suite reaches over 92\,\% line
coverage of the \code{loco\_mujoco} package, measured with
\code{coverage.py}. The measurement excludes code that cannot run in an
automated, offline setting: the interactive OpenGL viewer, the environments
that require the license-restricted MyoSkeleton model, the entry points that
download datasets or read the license-restricted SMPL model files, and the
package \code{\_\_init\_\_} files; the exclusions are listed with their
rationale in \code{pyproject.toml}. Tests that would download motion data use
synthetic trajectories instead, so the suite needs no network access.

A GitHub Actions workflow installs the package with its optional
extras and runs the suite with coverage on every push and pull request to the
\code{master} and \code{dev} branches, except for the task-factory tests, which
load the hosted datasets. The testing page of the documentation
(\url{https://loco-mujoco.readthedocs.io/en/latest/development/testing.html})
gives the commands that reproduce this run and the coverage measurement
locally.

\section{Simulation Throughput}
\label{app:throughput}

Table~\ref{tab:throughput} compares the three simulation backends of \lmj{}
with mjlab~\citep{zakka2026mjlab} and MuJoCo
Playground~\citep{zakka2025playground} on the Unitree~G1. Environment-level
throughput depends on the task as much as on the framework, so \lmj{} runs each
framework's own task: we rebuilt mjlab's flat-ground G1 velocity task and
Playground's G1 joystick task from \lmj{}'s components, on the model that each
framework compiles. A rebuilt task runs on the same model with the same MuJoCo
settings as the original; a step takes the same action and returns
the same observations, reward, termination, and command, and
resets the same way. Only the implementation differs. Replaying recorded steps
of the original tasks with their physics results injected, the rebuilt tasks
reproduce every observation, reward term, and termination to float32 rounding
on all three backends.
The table has four blocks, and numbers compare only within a block.

\paragraph{Blocks A and B: mjlab's task.}
mjlab's G1 has 34 collision geometries with self-collision and uses ten solver
and 20 line-search iterations. Block~A keeps full self-collision; block~B
keeps only the contacts between the 14 foot capsules and the floor (mjlab's
\code{FEET\_ONLY\_COLLISION}). The task runs without mjlab's domain
randomization events. On MuJoCo~Warp both frameworks get 48 contacts and 300
constraints per environment.

\paragraph{Block C: MuJoCo Playground's task.}
Playground's G1 has five explicit contact pairs and uses three solver and five
line-search iterations. The task runs with Playground's observation noise and
pushes. On MuJoCo~Warp both frameworks get 16 contacts and 128 constraints per
environment, because the shipped 8 and 90 are exactly full when the robot
stands.

\paragraph{Block D: \lmj{} alone.}
\lmj{}'s legacy G1 runs its own task with MJX-tuned settings: eight foot--floor
contacts, two solver and four line-search iterations, 100\,Hz.

\paragraph{Protocol.}
Blocks~A to~C use MuJoCo~3.11 and the Euler integrator (mjlab's default is
\code{implicitfast}) at 50\,Hz with four physics substeps of 5\,ms. We follow
mjlab's benchmark protocol: zero actions, 50 warm-up and 200 timed environment
steps (2{,}000 on the CPU). Each entry is the mean of five measurements in
each of two passes run in opposite order, which agree within 2.3\,\% from
1{,}024 environments on and within 3\,\% otherwise. The CPU backend runs one
environment per process. The measurements cover a single embodiment on a
single GPU, one NVIDIA RTX~4090 (24\,GB) with an Intel Core i9-13900KF.
mjlab~1.6.1 uses MuJoCo~Warp~3.11.0, Warp~1.19.0, and PyTorch~2.14.1;
\lmj{}~1.2.1 uses MJX~3.11.0, which bundles its own copy of MuJoCo~Warp, with
Warp~1.14.0, JAX~0.10.1, and \code{loco-mujoco-real}~0.1.0; MuJoCo
Playground~0.2.0, its JAX release, uses the same MJX, Warp, and JAX versions.
All GPU libraries are their CUDA~12 builds.

\begin{table}[t]
\centering\footnotesize
\setlength{\tabcolsep}{6pt}
\renewcommand{\arraystretch}{1.05}
\begin{tabular}{llrrrrr}
\toprule
 & & \multicolumn{5}{c}{parallel environments}\\
\cmidrule(lr){3-7}
framework & backend & 1 & 8 & 1{,}024 & 4{,}096 & 8{,}192\\
\midrule
\multicolumn{7}{l}{\textbf{A: \lmj{} and mjlab}, mjlab's G1 velocity task, full self-collision}\\
mjlab  & MjWarp       & 212     & ---      & 135{,}764 & 314{,}703 & \textbf{393{,}183} \\
\lmj{} & MjWarp       & 402     & ---      & \textbf{185{,}070} & \textbf{338{,}543} & 377{,}642 \\
\lmj{} & MJX          & 148     & ---      & 10{,}649  & 7{,}870   & 7{,}574 \\
\lmj{} & MuJoCo (CPU) & \textbf{1{,}126} & \textbf{8{,}296}  & ---       & ---       & --- \\
\addlinespace
\midrule
\multicolumn{7}{l}{\textbf{B: \lmj{} and mjlab}, mjlab's G1 velocity task, foot--floor contacts only}\\
mjlab  & MjWarp       & 213     & ---      & 138{,}262 & 322{,}327 & \textbf{403{,}180} \\
\lmj{} & MjWarp       & 406     & ---      & \textbf{190{,}612} & \textbf{352{,}790} & 394{,}745 \\
\lmj{} & MJX          & 172     & ---      & 52{,}684  & 68{,}343  & 71{,}557 \\
\lmj{} & MuJoCo (CPU) & \textbf{1{,}142} & \textbf{8{,}623}  & ---       & ---       & --- \\
\addlinespace
\midrule
\multicolumn{7}{l}{\textbf{C: \lmj{} and MuJoCo Playground}, Playground's G1 joystick task}\\
Playground & MjWarp   & 553     & ---      & 307{,}697 & 582{,}519 & 654{,}222 \\
\lmj{} & MjWarp       & 580     & ---      & \textbf{323{,}233} & \textbf{644{,}162} & \textbf{727{,}907} \\
Playground & MJX      & 267     & ---      & 128{,}130 & 198{,}491 & 217{,}588 \\
\lmj{} & MJX          & 271     & ---      & 130{,}834 & 201{,}070 & 220{,}322 \\
\lmj{} & MuJoCo (CPU) & \textbf{1{,}957} & \textbf{14{,}263} & ---       & ---       & --- \\
\addlinespace
\midrule
\multicolumn{7}{l}{\textbf{D: \lmj{} alone}, legacy G1 with its own task and MJX-optimized settings}\\
\lmj{} & MjWarp       & \textbf{710}     & ---      & \textbf{427{,}123} & \textbf{786{,}875} & \textbf{831{,}463} \\
\lmj{} & MJX          & 385     & ---      & 232{,}238 & 341{,}540 & 325{,}174 \\
\bottomrule
\end{tabular}
\caption{Environment steps per second on the Unitree~G1 by number of parallel
environments. In blocks A to C both frameworks run the same task on the same
model. Numbers compare only within a block; the fastest entry of each column
is bold within its block. A dash marks a configuration that was not measured.}
\label{tab:throughput}
\end{table}

\paragraph{Backends of \lmj{}.}
With full self-collision (block~A), MjWarp is 17 to $50\times$ faster than MJX, which
evaluates all 502 candidate geometry pairs of this robot in every step, whereas
MjWarp's broad phase discards distant ones. Restricting the contacts to the
feet (block~B), which leaves 14 pairs, makes MJX 4.9 to $9.4\times$ faster, but changes
MjWarp, mjlab, and the CPU backend by less than 5\,\%. With a single
environment the GPU backends are latency-bound and the CPU backend is the
fastest; with eight processes it scales 7.3 to $7.6\times$. MJX can be sped up
considerably with MJX-specific settings and very simple collision models: the
legacy G1 of \lmj{} reaches $3.4\times10^{5}$ environment steps per second with
MJX (block~D).

\paragraph{\lmj{} and mjlab.}
Both frameworks use MjWarp here. On mjlab's task (blocks~A and~B), \lmj{} is
1.36 to $1.38\times$ faster than mjlab at 1{,}024 environments, 1.08 to
$1.09\times$ faster at 4{,}096, and reaches 0.96 to $0.98\times$ mjlab's
throughput at 8{,}192. \lmj{} does extra work on this task: it
computes the observation of a reset environment separately, from its
kinematics. As shipped, with its domain randomization and
\code{implicitfast}, mjlab reaches 98{,}000, 244{,}000, and 313{,}000
environment steps per second at 1{,}024, 4{,}096, and 8{,}192 environments.

\paragraph{\lmj{} and MuJoCo Playground.}
On Playground's task (block~C), \lmj{} is 1.05 to $1.11\times$ faster than
Playground with MjWarp and 1.01 to $1.02\times$ with MJX. Here too \lmj{}
does extra work: at an automatic reset Playground restores a cached
initial state, whereas \lmj{} samples a new one.

\paragraph{GPU memory.}
Table~\ref{tab:memory} reports the peak GPU memory of the same configurations,
read with \code{nvidia-smi} for the process tree of a run of at least 250
environment steps, including the CUDA context of about 0.6\,GiB. The JAX frameworks donate
the state to their jitted step, so that its buffers are reused as in a compiled
training loop; mjlab updates its state in place. On mjlab's task, \lmj{}'s
MjWarp backend needs 1.4 to $1.8\times$ mjlab's memory, for the temporary
buffers of JAX's functional step; MJX with full self-collision needs far more,
16.6\,GiB at 8{,}192 environments. On Playground's task, \lmj{} needs as much as
Playground at 1{,}024 environments and 29 to $42\,\%$ less at 4{,}096 and
8{,}192.

\begin{table}[t]
\centering\footnotesize
\setlength{\tabcolsep}{6pt}
\renewcommand{\arraystretch}{1.05}
\begin{tabular}{llrrr}
\toprule
 & & \multicolumn{3}{c}{parallel environments}\\
\cmidrule(lr){3-5}
framework & backend & 1{,}024 & 4{,}096 & 8{,}192\\
\midrule
\multicolumn{5}{l}{\textbf{A: \lmj{} and mjlab}, full self-collision}\\
mjlab  & MjWarp     & \textbf{0.71} & \textbf{1.19} & \textbf{1.82} \\
\lmj{} & MjWarp     & 0.98 & 1.84 & 3.06 \\
\lmj{} & MJX        & 2.60 & 8.60 & 16.60 \\
\addlinespace
\midrule
\multicolumn{5}{l}{\textbf{B: \lmj{} and mjlab}, foot--floor contacts only}\\
mjlab  & MjWarp     & \textbf{0.68} & \textbf{1.10} & \textbf{1.60} \\
\lmj{} & MjWarp     & 0.98 & 1.84 & 2.94 \\
\lmj{} & MJX        & 0.71 & 1.46 & 2.59 \\
\addlinespace
\midrule
\multicolumn{5}{l}{\textbf{C: \lmj{} and MuJoCo Playground}}\\
Playground & MjWarp & 0.87 & 1.75 & 2.99 \\
\lmj{} & MjWarp     & 0.87 & 1.25 & 1.96 \\
Playground & MJX    & \textbf{0.70} & 1.45 & 2.48 \\
\lmj{} & MJX        & \textbf{0.70} & \textbf{0.95} & \textbf{1.45} \\
\bottomrule
\end{tabular}
\caption{Peak GPU memory in GiB of the GPU configurations of blocks A to C of
Table~\ref{tab:throughput}, with the same buffers, including the CUDA context.
The lowest entry of each column is bold within its block. The JAX frameworks
(\lmj{} and Playground) donate the state to their jitted step.}
\label{tab:memory}
\end{table}

\section{Experimental Details}
\label{app:experiments}
\setcounter{topnumber}{1}\setcounter{totalnumber}{1}%
\renewcommand{\floatpagefraction}{0.99}%

This appendix reports the experiments run with \lmj{}. Experiments~1
to~3 are simulation benchmarks on the same 13 clips of the native \lmj{}
dataset: experiment~1 compares embodiments, experiments~2 and~3 compare
algorithms on one humanoid. Experiment~4 trains one multi-clip policy per
robot on the real-robot environments of \code{loco-mujoco-real} and
shows the deployment on hardware.

\subsection{Experiment 1: PPO\,+\,DeepMimic across all humanoids}
\label{app:exp1}
\begin{figure}[t]
\centering
\includegraphics[width=.85\textwidth]{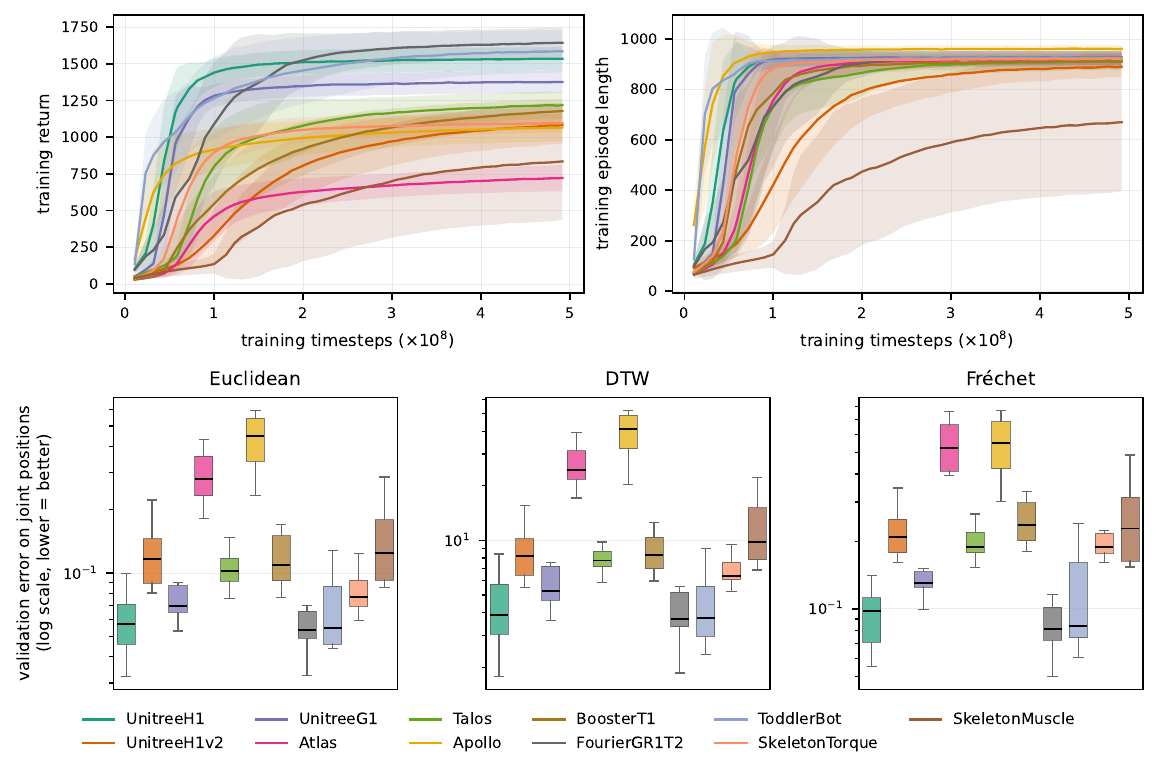}
\caption{PPO\,+\,DeepMimic on 11 embodiments, 13 clips, and 5 seeds.
\textbf{Top:} aggregate training return and episode length per
embodiment (mean $\pm$std). \textbf{Bottom:} final joint-position
Euclidean, DTW, and Fr\'echet errors per embodiment (log scale;
lower is better).}
\label{fig:exp1}
\end{figure}

The headline experiment of Figure~\ref{fig:exp1}: per-clip specialists
trained on every (env, dataset) cell, 11 embodiments, 13 clips, and
5 seeds each.

\begin{figure}[t]
\centering
\includegraphics[width=.7\textwidth]{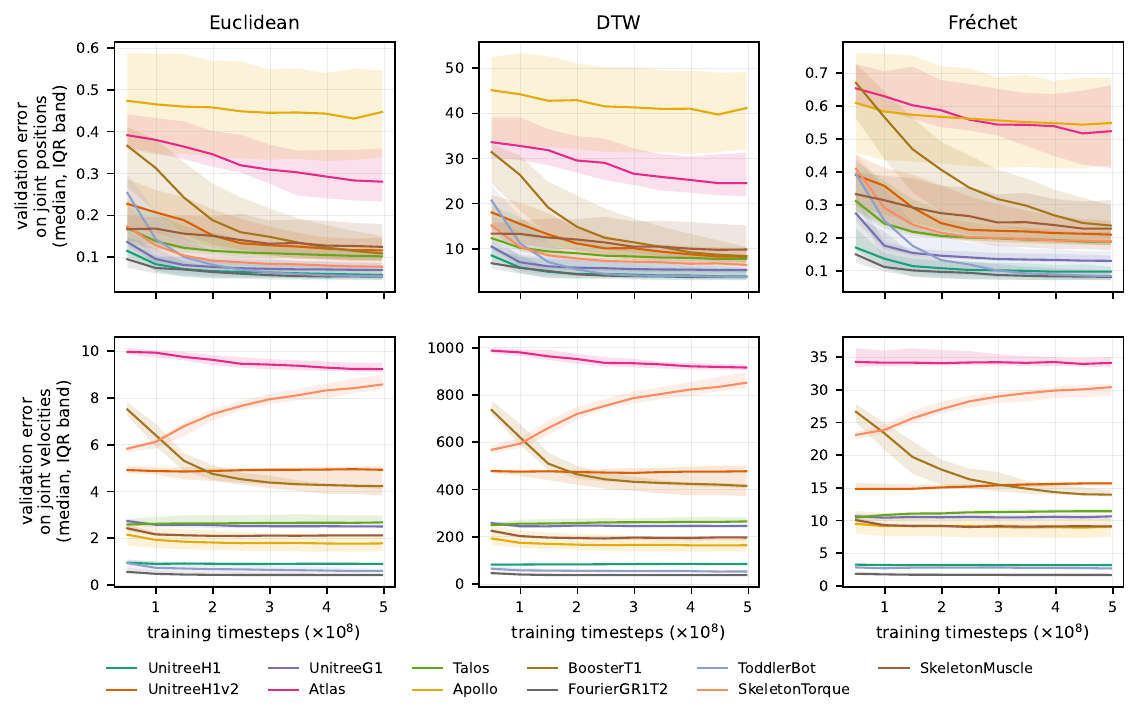}
\caption{Per-env validation error over training, with rows for joint
positions (\code{qpos}, mostly radians) and joint velocities
(\code{qvel}); columns are the three
trajectory-similarity measures. Each line is the median across the
5 seeds and 13 mocap clips contributing to each (env, time) point;
bands are the inter-quartile range.}
\label{fig:exp1-val-per-env}
\end{figure}

\paragraph{Trajectory-similarity measures.}
Return and episode length say how long a policy survives, not how
closely it follows the reference motion. \lmj{} therefore compares each
evaluation rollout with the reference clip it tracks, as two time series
of one kinematic quantity, using three measures from the \code{metrx}
library~\citep{pompetzki2024metrx}. The \emph{Euclidean} measure is the
mean Euclidean distance between rollout and reference at the same time
step, so it also penalizes a motion that is correct but early or late.
\emph{Dynamic time warping}~\citep{sakoe1978dtw} first aligns the two
series with a monotone time warp and sums the distances along the best
alignment, which makes it insensitive to such timing offsets. The
\emph{discrete Fr\'echet distance}~\citep{eiter1994frechet} uses the same
kind of alignment but takes the largest distance along it instead of
the sum, so it reports the worst deviation of the rollout rather than
the average one. All three are JIT-compiled, evaluated in parallel over
the evaluation environments, and averaged over rollouts; they are logged
during training next to the return (Figure~\ref{fig:exp1-val-per-env})
and reported at the end of training (Figure~\ref{fig:exp1}, bottom).

\begin{figure}[t]
\centering
\includegraphics[width=.7\textwidth]{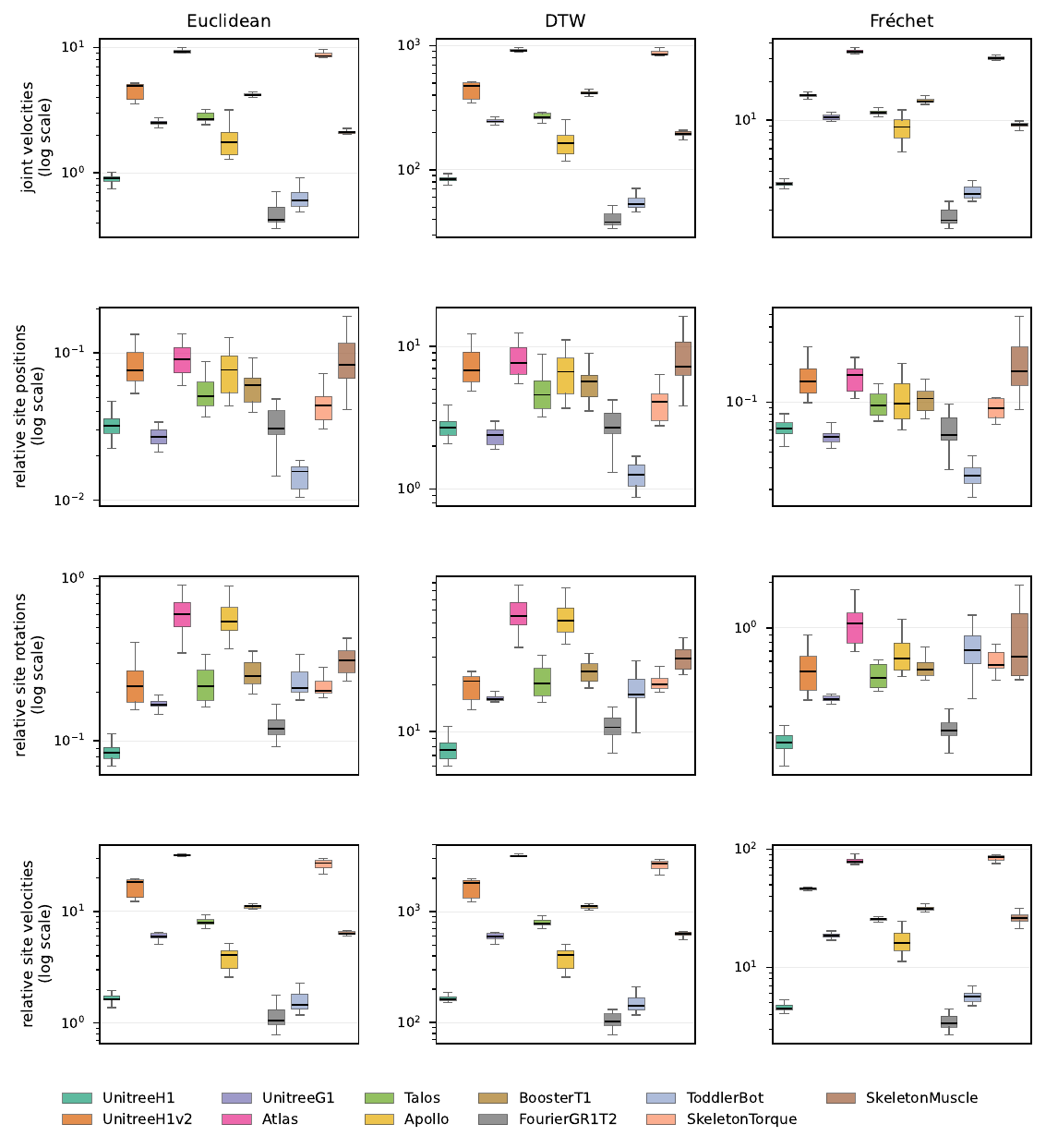}
\caption{Per-env distribution of the final validation error on four
further quantities (rows) under the three trajectory-similarity
measures (columns), complementing the joint-position panel of
Figure~\ref{fig:exp1}. The final policies are evaluated again after
training, each with five independent validation rollouts that are
averaged. Each box aggregates 5 seeds $\times$ 13 mocap
clips for one env (whiskers at 1.5$\times$IQR, log $y$).}
\label{fig:exp1-app-other-quantities}
\end{figure}

\paragraph{Validation quantities.} The trajectory-similarity measures
are evaluated here on five of the 11 supported per-step quantities. Joint positions
(\code{qpos}, shown in Figure~\ref{fig:exp1}) and joint velocities
(\code{qvel}) are the raw configuration-space states. The remaining
three are computed at a small set of \emph{mimic sites} — markers
welded to body links (head, hands, feet, pelvis, etc.) that mirror
the SMPL-joint correspondences used by the retargeting pipeline
(Appendix~\ref{app:retargeting}). Their world poses are expressed
in the root frame of the robot, giving root-relative site positions
(\code{site\_rpos}), root-relative orientations
(\code{site\_rrotvec}, in axis-angle), and root-relative linear
velocities (\code{site\_rvel}). The relative-site quantities are
the dominant signal in the DeepMimic reward used for training: by
construction the policy is rewarded for matching end-effector
trajectories \emph{relative to the body}, which is what makes the
imitation transfer across embodiments with different proportions.
Figure~\ref{fig:exp1-app-other-quantities} extends the validation panels
of Figure~\ref{fig:exp1} to \code{qvel}, \code{site\_rpos},
\code{site\_rrotvec}, and \code{site\_rvel}; the two joint-space
quantities are shown over training in
Figure~\ref{fig:exp1-val-per-env}.

\paragraph{Final error per embodiment.}
In Figure~\ref{fig:exp1-app-other-quantities}, the relative-site rows are
the quantities that drive the DeepMimic reward signal during training.
Within every row the three measures order the embodiments almost
identically (the main exception is the Fr\'echet distance on the site
rotations, under which ToddlerBot ranks markedly worse), supporting the observation that they rank converged
PPO\,+\,DeepMimic policies similarly. Across rows the ordering on the
site positions and site rotations is close to that of the
joint-position panel of Figure~\ref{fig:exp1} (Spearman
$\rho \approx 0.9$ between the per-env medians under the Euclidean
measure), while the two velocity quantities order the embodiments
differently ($\rho \approx 0.6$).

\paragraph{Error over training.}
Figure~\ref{fig:exp1-val-per-env} shows the validation error over
training. We use median/IQR rather than mean/CI because a small number
of failure-mode seeds has a final-step error 50--500$\times$ that of
the typical seed, which would otherwise dominate the mean. For each
quantity the per-env ordering is highly consistent across measures,
supporting the empirical observation that the three measures rank
converged PPO\,+\,DeepMimic policies similarly at the (env, dataset)
level.

\begin{figure}[t]
\centering
\includegraphics[width=.7\textwidth]{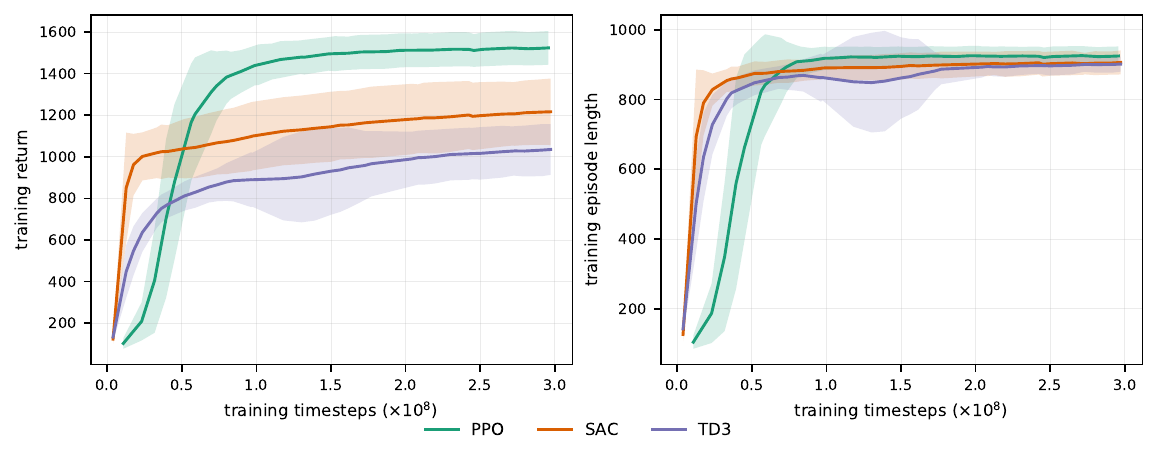}
\caption{Algorithm comparison on UnitreeH1 (experiment 2): PPO
vs.\ SAC vs.\ TD3 aggregated across all 13 native clips and 5
seeds (65 runs per algorithm). \textbf{Left:} training return.
\textbf{Right:} training episode length. Lines are
algorithm-level mean over (task, seed) at each binned step, bands
are $\pm$\,std.}
\label{fig:exp2-h1}
\end{figure}
\subsection{Experiment 2: PPO vs.\ SAC vs.\ TD3 on UnitreeH1}
\label{app:exp2}
An algorithm-side ablation on a single humanoid, isolating the
on-policy / off-policy / deterministic distinction from embodiment
confounds.
We trained the three algorithms on UnitreeH1 across all 13 native
clips (\code{walk}, \code{run}, \code{squat},
\code{stepinplace1}--\code{3}, \code{onestepleft},
\code{onestepright}, \code{onestepside1}--\code{2},
\code{onesteplong}, \code{onehopforward}, \code{highknee}), 5
seeds per cell. Same hyper-parameter file, same DeepMimic reward,
same 1000-step horizon. Figure~\ref{fig:exp2-h1} aggregates all
65 runs per algorithm into a single mean\,$\pm$\,std curve over
training. PPO reaches the highest return ($\sim$1530), SAC follows
at $\sim$1260, and TD3 trails at $\sim$1080; episode lengths
converge to $\sim$900--925 for all three. This result is one reason
\lmj{} ships PPO\,+\,DeepMimic as the default training recipe.

\subsection{Experiment 3: PPO vs.\ GAIL on UnitreeH1}
\label{app:exp3}
\begin{figure}[t]
\centering
\includegraphics[width=.75\textwidth]{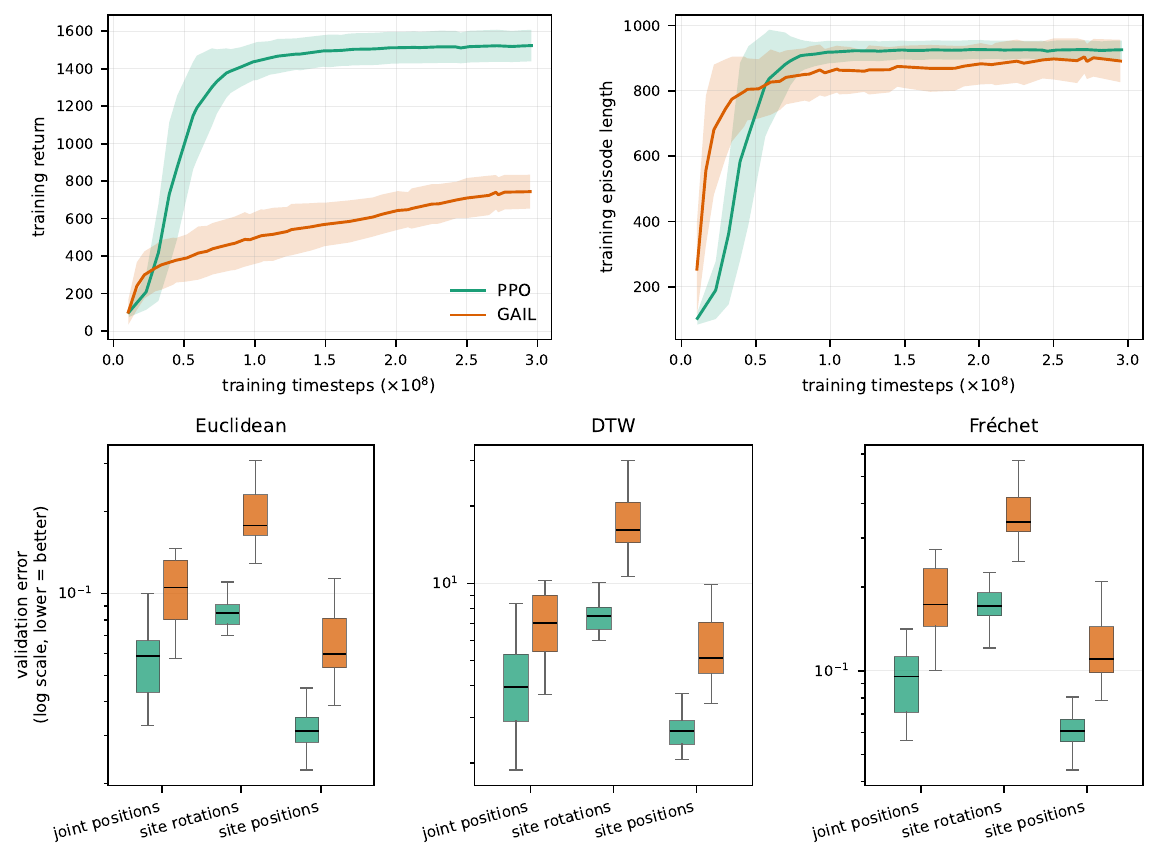}
\caption{PPO vs.\ GAIL on UnitreeH1 (experiment 3), aggregated
across 12 native clips (\code{onestepside1} excluded) and 5 seeds
per algorithm.
\textbf{Top:} training return and training episode length, mean
$\pm$\,std band over (task, seed) at each binned step (capped at
$3\!\times\!10^{8}$ steps).
\textbf{Bottom:} per-algorithm distribution of the final-step
validation error, one panel per trajectory-similarity measure
(Euclidean / DTW / Fr\'echet); within each panel, PPO and GAIL
boxes are paired across three kinematic quantities — joint
positions (\code{qpos}), site rotations (\code{site\_rrotvec},
axis-angle), and site positions (\code{site\_rpos}). Each box
aggregates 5 seeds $\times$ 12 clips for one algorithm (whiskers
at 1.5$\times$IQR, log $y$).}
\label{fig:exp3-h1}
\end{figure}
We compare on-policy DeepMimic (PPO) against adversarial imitation
(GAIL) on the same UnitreeH1 environment. We use 12 of the 13
clips of experiment~2 (\code{onestepside1} is dropped
because the GAIL batch was one seed short on that clip), 5 seeds
per cell. Same hyper-parameter file, same 1000-step horizon; the
only training-time difference is whether the reward is the
DeepMimic mocap-tracking signal (PPO) or the discriminator output
(GAIL). Figure~\ref{fig:exp3-h1} mirrors the layout of
Figure~\ref{fig:exp1}, but the box plots compare the two
algorithms instead of the 11 embodiments. PPO converges to a
$\sim$2$\times$ higher training return and a lower validation error
on all three quantities under all three trajectory-similarity
measures: the median error is about 1.8$\times$ lower on joint
positions and 1.8 to 2.2$\times$ lower on site positions and
rotations. As in Figure~\ref{fig:exp1-app-other-quantities}, the
errors are computed from the final policies with five validation
rollouts each. The gap is consistent with the
well-known difficulty of training a discriminator on
high-dimensional whole-body humanoid trajectories without an
explicit per-step shaping signal; \lmj{} ships GAIL primarily
as a comparison baseline and a starting point for AMP-style
hybrids.

\subsection{Experiment 4: SuperMimic on the real-robot environments}
\label{app:exp4}
\setcounter{topnumber}{2}\setcounter{totalnumber}{2}%
Table~\ref{tab:supermimic} reports SuperMimic on the two real-robot
environments of \code{loco-mujoco-real}, in simulation. For each robot, one policy is
trained on all 40 LAFAN1 clips for $1.53\times10^{8}$ environment steps
with 4{,}096 parallel MjWarp environments, domain randomization, pushes,
and self-collisions, which takes about 50~minutes on one RTX~6000~Ada.
The trained policy then plays every clip once from its first frame,
without noise or randomization; a clip counts as failed if the rollout
terminates before its end.

\paragraph{Domain randomization for sim-to-real transfer.}
Because these policies are meant to run on hardware, training relies
heavily on domain randomization, all of it built from \lmj{}'s
domain-randomization plug-in and identical for the two robots. It
targets sensing and disturbances rather than the simulated plant:
(i)~random pushes, a velocity impulse on the root of up to
0.5\,m/s horizontally and 0.78\,rad/s in yaw every 1--3\,s;
(ii)~a per-episode encoder bias of up to $\pm0.02$\,rad on the measured
joint positions; (iii)~uniform noise on the measured joint positions
($\pm0.01$\,rad), joint velocities ($\pm0.5$\,rad/s), and base angular
velocity ($\pm0.2$\,rad/s); (iv)~a friction coefficient drawn from
$[0.3, 1.2]$ for all colliding robot geometries; and (v)~an offset of
the pelvis center of mass of up to 5\,cm per axis. Noise and bias are
applied only to the actor's observations, while the critic receives the
clean values. The evaluation in Table~\ref{tab:supermimic} switches all
of this off, so it measures tracking quality rather than robustness.

\begin{figure}[t]
\centering
\includegraphics[width=\textwidth]{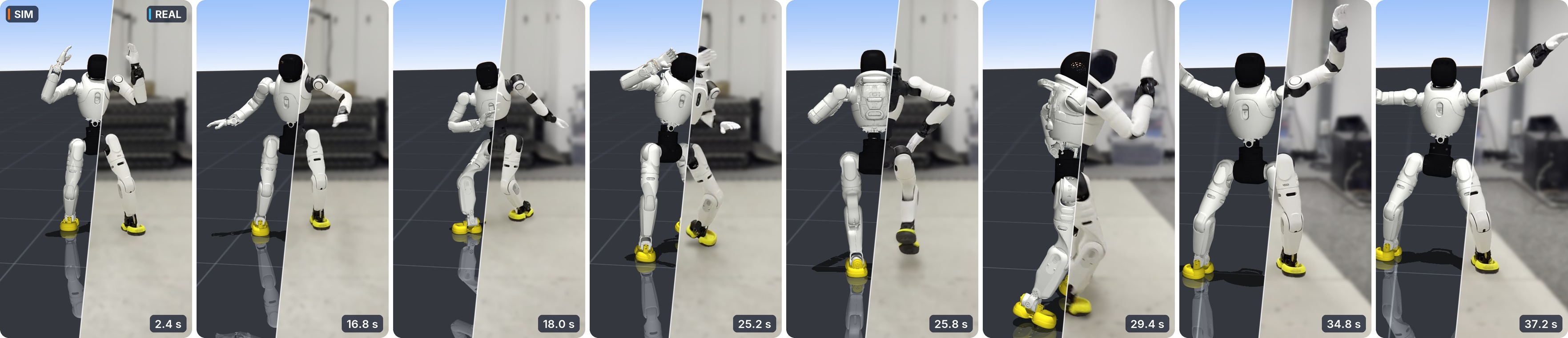}
\caption{Simulation and real robot for the motion of
Figure~\ref{fig:sim2real}. Each panel shows the simulation (left) and
the real AgiBot~X2~Ultra (right) at the same time of the motion.}
\label{fig:sim2real-split}
\end{figure}

\begin{table}[t]
\centering\small
\begin{tabular}{@{}lccc@{}}
\toprule
\textbf{Robot} & \textbf{Normalized return} & \textbf{Failed clips (of 40)} & \textbf{Completed frames} \\
\midrule
AgiBot~X2~Ultra & $0.776 \pm 0.004$ & 1,\,0,\,0 & 99.7\,\% \\
Unitree~G1      & $0.674 \pm 0.020$ & 2,\,0,\,1 & 99.1\,\% \\
\bottomrule
\end{tabular}
\caption{SuperMimic on all 40 LAFAN1 clips with one policy per robot,
trained for $1.53\times10^{8}$ environment steps.
The normalized return is the undiscounted reward sum divided by the
maximum attainable one, averaged over clips (mean $\pm$ standard
deviation over three seeds); failed clips are listed per seed, and
completed frames are averaged over seeds.}
\label{tab:supermimic}
\end{table}

\paragraph{From checkpoint to robot.}
\code{loco-mujoco-real} exports a checkpoint to a \emph{bundle}: the
network weights, the controller settings, and the reference motions,
which run with NumPy and MuJoCo alone. Three checks precede hardware. A
validation script confirms that the bundle reproduces the observations
and actions of the training environment term by term, and a rollout in
plain MuJoCo tests the export. In sim-to-sim, the controller and a
simulated robot then run as separate processes over the robot's own
middleware (AgiBot's AimDK on ROS~2, or the Unitree SDK), so the
controller already acts on delayed state.

\paragraph{Control loop.}
The policy runs at 50\,Hz. Its
observation uses onboard sensing only: joint positions and velocities
from the encoders, projected gravity and angular velocity from the
pelvis IMU, the previous action, and the reference frame to track.
There is no base-state estimate and no external tracking. Actions are joint-position targets, tracked by the drives' own
PD loop with the stiffness and damping of the simulated position servos.
On the X2~Ultra, a policy step takes about 2\,ms on state that is about
2\,ms old; the controller runs on a laptop over Ethernet or on the
robot's own computer, and damps the robot on stale state, a joint past
its limits, or a pelvis tilt beyond about $60^\circ$.

\paragraph{Hardware run.}
Figure~\ref{fig:sim2real-split} shows simulation and robot side by side
for the motion of Figure~\ref{fig:sim2real}, a 43\,s routine of chained
BONES-SEED~\citep{bones2026seed} dance clips. The policy is not one of
Table~\ref{tab:supermimic}: SuperMimic trained it in the same
environment on all of BONES-SEED, where it completes 99.9\,\% of the
clips in simulation after $4.56\times10^{8}$ environment steps.

\clearpage
\bibliography{bibliography}

\end{document}